\documentclass[lettersize,journal]{IEEEtran}
\usepackage{amsmath,amsfonts}
\usepackage{algorithmic}
\usepackage{algorithm}
\usepackage{array}
\usepackage[caption=false,font=normalsize,labelfont=sf,textfont=sf]{subfig}
\usepackage{textcomp}
\usepackage{stfloats}
\usepackage{url}
\usepackage{verbatim}
\usepackage{graphicx}
\usepackage{cite}
\usepackage{enumitem}
\usepackage{xcolor}
\usepackage{pifont}
\usepackage{multirow}
\usepackage{threeparttable}

\usepackage[colorlinks=true, linkcolor=blue, citecolor=blue, urlcolor=blue]{hyperref}
\begin{document}

\title{Continuous Cross-Domain Traffic State Prediction via Memory-Augmented Graph Liquid Time-Constant Networks}

\author{Jinrong Xiang and Ming Xu$^{*}$%
\thanks{$^{*}$Corresponding author: Ming Xu.}%
\thanks{Jinrong Xiang and Ming Xu are with the Software College, Liaoning Technical University, Huludao, Liaoning 125100, China 
(e-mail: xiangjinrong2025@163.com; xum.2016@tsinghua.org.cn).}%
}



\maketitle

\begin{abstract}
Traffic state prediction is a fundamental task in intelligent transportation systems. In practical applications, some regions suffer from limited traffic observations due to insufficient sensing infrastructure, making cross-domain knowledge transfer an important solution for data-scarce traffic prediction. However, existing cross-domain traffic prediction methods still face several limitations, including coarse-grained source-target adaptation, limited capability in handling unseen target-domain patterns, and insufficient modeling of continuous traffic dynamics under irregular or heterogeneous temporal conditions. To address these issues, this paper proposes a continuous cross-domain traffic prediction framework, termed Memory-Augmented Graph Liquid Time-Constant Network (MA-GLTC). Specifically, we first construct spatio-temporal units (STUs) to decompose traffic networks into transferable local units, enabling fine-grained knowledge alignment across domains. Then, a graph liquid time-constant network (GLTC) is developed to model graph-coupled traffic evolution in continuous time. Different from generic graph neural ODE-based models, GLTC introduces graph-coupled recurrent conductance into liquid time-constant dynamics, allowing node states to evolve with leakage, adaptive time constants, and neighborhood-aware feedback. Furthermore, a Memory-based Transfer Storage (MTS) mechanism is designed to preserve source-domain knowledge, retrieve matched traffic patterns, and update reliable target-domain patterns when unseen states emerge. Experiments on five public traffic datasets demonstrate that MA-GLTC consistently outperforms representative inner-domain and cross-domain baselines in both short-term and long-term prediction tasks. Compared with the second-best method, MA-GLTC reduces the average prediction errors by 3.02\%, 0.33\%, 8.92\%, 10.09\%, and 2.11\%, respectively.
\end{abstract}

\begin{IEEEkeywords}
Traffic prediction, Cross-domain knowledge transfer, Liquid time-constant network, Continuous spatio-temporal modeling.
\end{IEEEkeywords}

\section{Introduction}
\IEEEPARstart{T}{raffic} prediction is a fundamental task in intelligent transportation systems (ITS) and plays a critical role in smart city management \cite{CGSTT}. Since urban traffic systems are naturally organized as road networks, traffic prediction is inherently a spatio-temporal graph modeling problem that requires capturing both spatial dependencies among road segments and temporal evolution patterns in traffic states. However, modeling these spatial and temporal dependencies usually relies on sufficient historical observations, which are not always available in real-world transportation scenarios. New or low-resource regions often suffer from limited sensor coverage and insufficient historical data, making it difficult to train reliable prediction models directly \cite{MTPB}. Moreover, traffic systems across regions may differ in road topology, sensing configurations, and mobility patterns, further increasing the difficulty of knowledge transfer. To address these problems, cross-domain graph learning provides a promising paradigm by transferring structural and temporal knowledge from data-rich source regions to data-scarce target regions, thereby improving traffic prediction performance in low-resource scenarios \cite{ref2,zang2024transfer,CCMHC}.

Although recent studies have integrated graph-based modeling with cross-domain transfer learning for traffic prediction, existing methods still have several limitations in addressing the above challenges.

\begin{figure*}[htbp]
    \centering
    \subfloat[]{
        \includegraphics[width=0.45\textwidth]{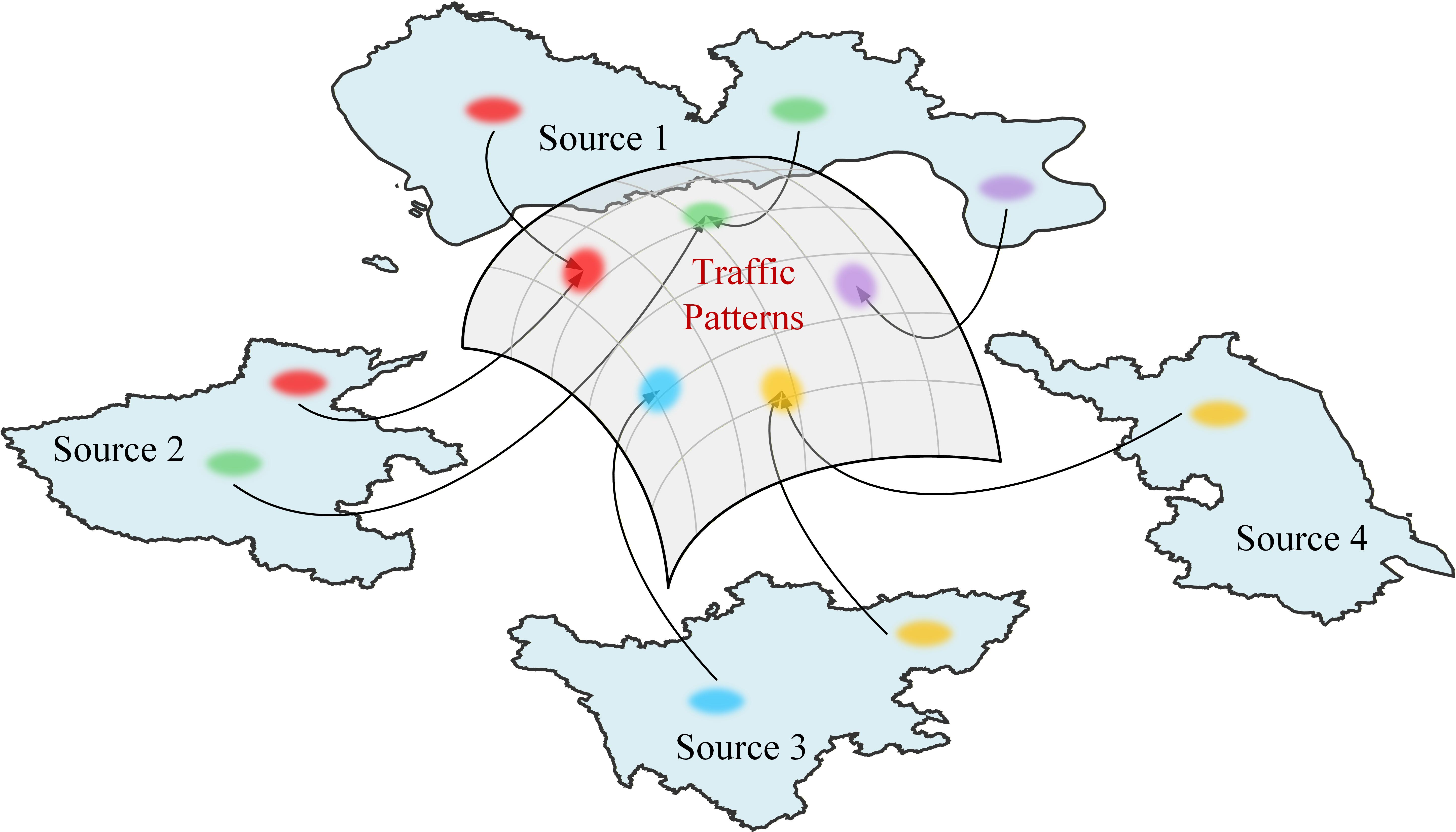}
        \label{fig:ms}
    }
    \hfill
    \subfloat[]{
        \includegraphics[width=0.45\textwidth]{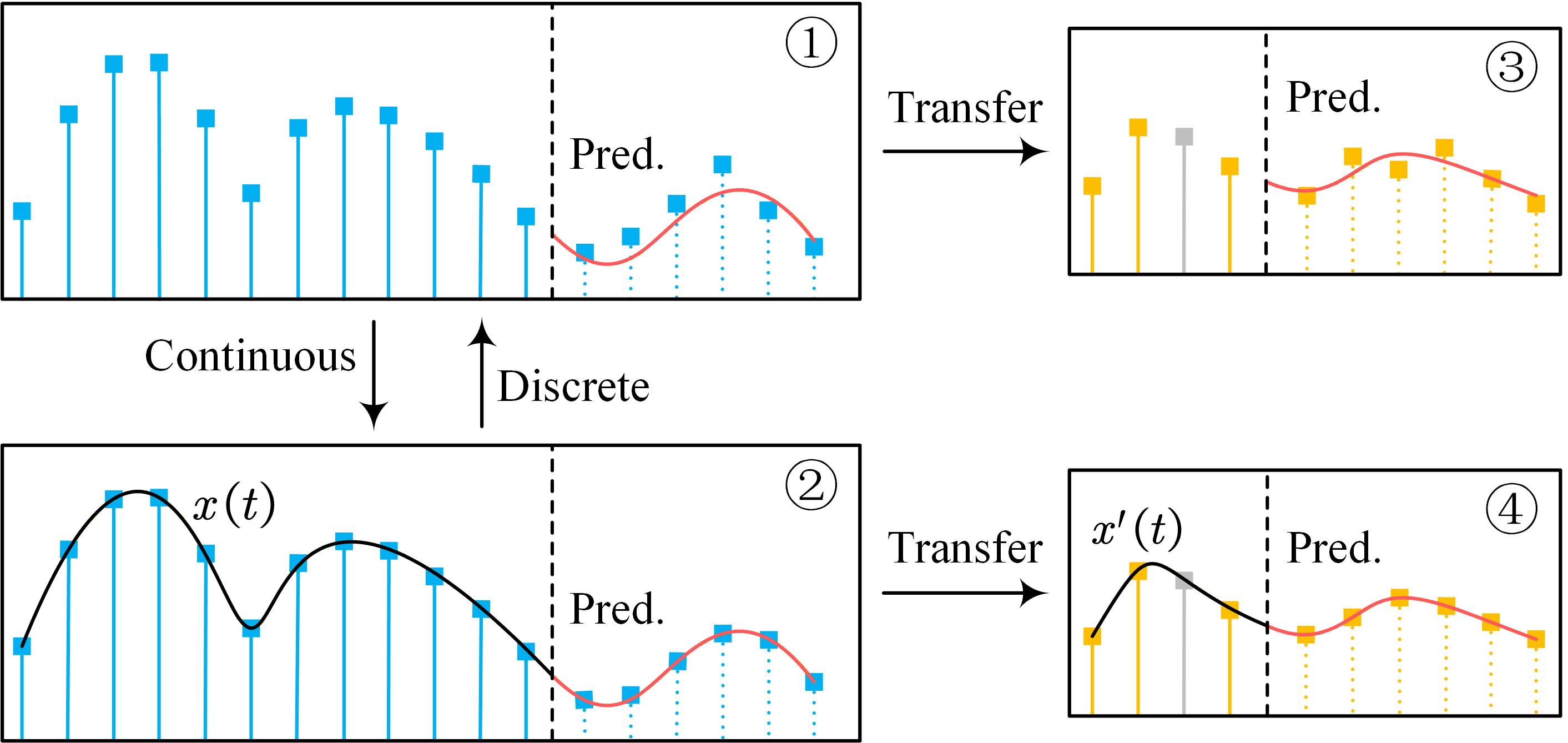}
        \label{fig:pt}
    }
    \caption{Illustration of key challenges in cross-domain traffic prediction.  
    (a) Limitations of cross-domain transfer paradigms. Single-source (e.g., Source 1) provides insufficient pattern coverage; multi-source enriches the space but introduces feature redundancy and information noise (e.g., overlapping red patterns from Source 1 and Source 2). 
    (b) Comparison of discrete and continuous temporal transfer. Gray squares indicate missing data. The black curve models continuous historical traffic dynamics, while the red curve represents the predicted continuous trajectory of future traffic flow.}
    \label{fig:motivation}
\end{figure*}

\begin{enumerate}[label=\roman*)]
    \item \textbf{Existing models lack graph-coupled structured continuous dynamics.} In cross-domain traffic prediction, source and target regions may differ in road topology, traffic evolution patterns, and sensing configurations. Therefore, the model is expected to capture traffic dynamics that are both transferable across domains and stable under domain shifts. Recently, neural differential equation-based models have been introduced into graph-based traffic forecasting, including graph neural ordinary differential equations (ODEs)~\cite{gde}, spatio-temporal graph neural controlled differential equations (CDEs)~\cite{stgncde}, graph neural rough differential equations (RDEs)~\cite{stgnrde}, and multi-ODE graph networks~\cite{gramode}. Although these methods provide continuous-time modeling capabilities, they usually rely on generic neural differential functions or multiple ODE parameterizations, where structured mechanisms such as leakage, input-dependent time constants, and conductance-based state transitions are not explicitly modeled. Liquid time-constant networks (LTCs) introduce adaptive time constants for continuous sequence modeling~\cite{LTc}, but standard LTCs are primarily formulated for vector-valued sequences and do not explicitly incorporate graph-structured neighborhood feedback without graph-aware adaptations. As a result, they are not directly suitable for modeling continuous traffic dynamics over road networks.

    \item \textbf{Existing transfer paradigms suffer from limited pattern coverage and insufficient continual adaptation} Cross-domain traffic prediction methods can generally be divided into single-source and multi-source transfer paradigms. As illustrated in Fig. \ref{fig:motivation}\subref{fig:ms}, single-source transfer methods are often constrained by limited traffic pattern diversity \cite{sd},  making it difficult to cover complex and heterogeneous target-domain dynamics. Multi-source transfer methods attempt to introduce richer traffic patterns from multiple source domains \cite{md}, but directly aggregating heterogeneous knowledge may lead to feature conflicts and negative transfer. Moreover, practical constraints such as privacy concerns, deployment costs, and data acquisition barriers may limit the availability of multiple source domains. More importantly, many existing methods are trained under relatively static transfer settings and provide limited mechanisms for continual adaptation. Consequently, they may become less effective when previously unseen traffic patterns gradually emerge in evolving target-domain environments.

    \item \textbf{Temporal granularity mismatch and irregular observations further complicate cross-domain transfer.} In practical ITS applications, source and target domains may be collected under different sensing infrastructures, resulting in inconsistent sampling intervals, missing observations, or irregular temporal records. Most existing graph-based traffic prediction models operate under a fixed discrete-time setting and implicitly assume that observations from different domains are temporally aligned. However, this assumption is difficult to satisfy in real-world transfer scenarios. As shown in Fig. \ref{fig:motivation}\subref{fig:pt}, directly transferring discrete traffic sequences across domains with different temporal granularities may cause temporal misalignment and degrade prediction performance. Therefore, a continuous-time modeling paradigm is needed to infer traffic states at arbitrary time points and better accommodate irregularly sampled target-domain observations.
\end{enumerate}

To address these challenges, this paper proposes a Memory-Augmented Graph Liquid Time-Constant Network (MA-GLTC) for continuous cross-domain traffic prediction. MA-GLTC consists of three key components: spatio-temporal unit (STU) decomposition, graph liquid time-constant network (GLTC), and memory-based transfer storage (MTS). Specifically, STU decomposition constructs fine-grained spatial and temporal units to facilitate localized source-target alignment. GLTC extends liquid time-constant dynamics to traffic graphs by incorporating leakage, adaptive time constants, and neighborhood-aware recurrent feedback, enabling structured continuous-time modeling of graph-coupled traffic evolution. MTS stores and updates transferable traffic patterns to support continual adaptation when unseen target-domain states emerge. By integrating these components, MA-GLTC provides a unified framework for handling source-target domain shifts, continuous traffic dynamics, and irregular temporal observations.

The main contributions of this work are summarized as follows.

\begin{enumerate}[label=\roman*)]
    \item We propose a knowledge transfer framework based on spatio-temporal unit decomposition for cross-domain traffic prediction. By decomposing traffic data into fine-grained spatio-temporal units, the proposed framework facilitates localized spatial transfer and parallel temporal pattern transfer, thereby improving knowledge alignment across spatial and temporal dimensions.
    \item We design a continuous-time prediction module, termed GLTC, to model graph-coupled traffic dynamics. Unlike existing graph neural ODE-based methods that mainly rely on generic neural differential functions, GLTC explicitly incorporates leakage, adaptive time constants, and neighborhood-aware recurrent feedback into liquid time-constant dynamics, enabling structured and stable continuous-time traffic state prediction.
    \item We develop an adaptive memory mechanism to preserve transferable traffic patterns and support continual adaptation. The memory module can be dynamically updated with emerging target-domain states, allowing the model to better accommodate evolving and previously unseen traffic patterns.
\end{enumerate}

In addition, owing to its continuous-time formulation, the proposed MA-GLTC framework can infer traffic states at arbitrary time points, reducing its dependence on fixed sampling intervals and improving robustness under temporal granularity mismatch, missing observations, and irregularly sampled target-domain data.

The rest of this paper is organized as follows. The related work and the proposed MA-GLTC are described in Sections II and III, respectively. Section IV demonstrates the computational experiments and results, and Section V summarizes this study and discusses the future work.

\section{RELATED WORK}
This section reviews two lines of studies: traffic prediction and liquid time-constant networks.

\subsection{Traffic prediction}
Recent years have seen rapid progress in traffic prediction, especially in spatio-temporal modeling and cross-domain generalization. Related studies can be broadly divided into graph-based and cross-domain traffic prediction. 

Graph-based methods represent traffic networks as graphs and learn spatio-temporal dependencies from observations. For spatial modeling, existing methods mainly include GNN-based models, self-attention-based models, and hybrid architectures~\cite{ref7}. GNN-based methods capture spatial correlations with predefined or learned graphs, but are often limited by topology and message-passing depth~\cite{ref8,ref9,ref10}. Self-attention-based methods offer more flexible dependency modeling but usually incur higher computational costs~\cite{DSTAN}. To combine structural inductive bias and dynamic dependency learning, several studies further integrate self-attention mechanisms into graph-based architectures~\cite{PatchSTG,MGCN,DAGCAN}. For temporal modeling, RNNs~\cite{ref12,ref13}, TCNs, and Transformers are widely used. However, these methods generally rely on sufficient labeled data in the target region, limiting their applicability to data-scarce scenarios. 

To address this issue, cross-domain traffic prediction has attracted increasing attention. Existing studies mainly aim to transfer useful knowledge from source regions to target regions by reducing domain discrepancies. Some methods focus on spatial adaptation, such as spatial homogeneity-aware transfer frameworks~\cite{shtl} and graph matching-based methods~\cite{ref16}, which attempt to align structural differences between source and target traffic networks. However, these methods often underexploit temporal distribution shifts across regions. Other studies emphasize temporal transfer, such as GLA-DA~\cite{gla}, which performs global-local alignment for multivariate traffic series, but pays limited attention to spatial structural knowledge. Recent works further explore joint spatio-temporal transfer. For example, CGSTT~\cite{CGSTT} conducts bilevel adaptation through spatial graph matching and temporal graph embedding, while ST-LLM+~\cite{llm} introduces large language models to enhance dependency modeling with network topology information. 

Despite these advances, existing cross-domain traffic prediction methods still face challenges in graph-coupled structured continuous dynamics, pattern coverage and continual adaptation, as well as temporal granularity mismatch and irregular observations. These limitations motivate us to develop a continuous cross-domain traffic prediction framework with structured dynamic modeling and memory-augmented adaptation capability.

\subsection{Overview of LTC Networks}

spatio-temporal graph neural networks (STGNNs) have been widely used in traffic prediction by jointly modeling road network topology and temporal dependencies~\cite{ref17,ref18,ref19}. Most existing STGNNs rely on discrete-time recurrent units, temporal convolutions, or attention mechanisms to capture temporal dynamics. Although effective, these modules usually describe temporal evolution through stacked discrete operations, making it difficult to explicitly characterize continuous state evolution and stability over time.

To overcome this limitation, neural ordinary differential equation (ODE)-based models~\cite{ode} have been introduced to formulate hidden representations as continuous-time dynamical systems. For example, recent studies model spatio-temporal representations as an initial value problem and update hidden states through numerical integration~\cite{odest}. These methods provide continuous-time modeling ability, but their dynamics are usually parameterized in a generic form. As a result, important structural priors, such as leakage, forgetting, and adaptive time constants, are not explicitly encoded. In addition, their computational cost and training stability are often sensitive to the choice of numerical solvers.

Liquid time-constant networks (LTCs) provide a more structured alternative for continuous-time sequence modeling. As a type of continuous-time recurrent neural network, LTCs introduce leakage terms and input-dependent gates into the hidden-state dynamics, allowing the effective time constants to change according to the input state. This mechanism enables LTCs to model non-stationary temporal evolution while maintaining more stable state transitions. Therefore, LTCs are particularly suitable for traffic prediction scenarios, where traffic states evolve continuously and are affected by time-varying patterns such as peak hours, incidents, and demand fluctuations.

Inspired by these properties, this work adopts LTC as the temporal dynamics backbone and further extends it to graph-structured traffic data. By coupling LTC-based temporal evolution with graph message passing, the proposed model can capture continuous traffic dynamics while preserving spatial dependencies among sensors.

\begin{table*}[!t]
\caption{Notations and their descriptions}
\label{tab:Notation}
\centering
\small
\setlength{\tabcolsep}{10pt}
\renewcommand{\arraystretch}{1.2}
\begin{tabular}{m{.13\textwidth} m{.72\textwidth}}
\hline\hline
\textbf{Notation} & \textbf{Description} \\
\hline
$G=(\mathcal{V},\mathcal{A})$ & Regional traffic graph with a set of $N$ sensor nodes $\mathcal{V}$ and a spatial adjacency matrix $\mathcal{A}\in\mathbb{R}^{N\times N}$. \\
$\mathcal{X}\in\mathbb{R}^{\tau_\mathrm{h} \times N\times D}$ & Historical input traffic graph signal tensor, where $D$ denotes the number of traffic features and $\tau_{\mathrm{h}}$ denotes the length of the historical input window. \\
$\mathcal{Y}\in\mathbb{R}^{\tau_\mathrm{f} \times N\times D}, \hat{\mathcal{Y}}$ & Future ground-truth and predicted traffic graph signal tensors, where $\tau_{\mathrm{f}}$ denotes the prediction horizon. \\
$\mathcal{U}_{m,k}$ & Spatio-temporal unit (STU) composed of the temporal unit $T_m$ and the spatial unit $\mathcal{V}_{m,k}$. \\
$\mathcal{S} , \mathcal{T}$ & Source domain with abundant data and target domain with limited data, respectively. \\
$\mathcal{L}_\mathrm{sc}, \mathcal{L}_\mathrm{ft}$ & Loss functions for source training stage and fine-tuning stage. \\
$\psi_{\mathcal{S}}^{(m,k)}$  
& STU-specific decoder parameters learned for the source-domain STU $\mathcal{U}_{m,k}^{\mathcal{S}}$. \\
$\psi_{\mathcal{T}}^{(m,k)}$  
& Decoder parameters adapted for the target-domain STU $\mathcal{U}_{m,k}^{\mathcal{T}}$. \\
$\mathcal{H}$ & Hidden-layer state. \\
$\mathcal{M}$ & Node validity mask. \\
$\mathbf{t}_{\Delta}$  
& Temporal feature vector for the future prediction step $\Delta$. \\
$\mathcal{Z}^{(m,k)}$ 
& Latent node representation generated from $\mathcal{H}^{(m,k)}$. \\
$\mathbf{P}_{\Delta}$ 
& Query-specific projection matrix generated from $\mathbf{t}_{\Delta}$. \\
$\mathcal{B}^{(m,k)}_{:,\Delta}$ 
& Trend baseline of STU $\mathcal{U}_{m,k}$ at the queried future horizon $\Delta$. \\
$\hat{\mathcal{R}}^{(m,k)}_{:,\Delta}$ 
& Residual prediction of STU $\mathcal{U}_{m,k}$ at the queried future horizon $\Delta$. \\
$CMD_P$ & $P$-order central moment discrepancy (CMD) metric. \\
\hline\hline
\end{tabular}
\end{table*}

\section{METHODOLOGY}
This section presents the design of MA-GLTC. We first introduce the key definitions and formulate the cross-domain traffic prediction problem, followed by an overview of the proposed framework. The main notations used in this paper are summarized in Table \ref{tab:Notation} for ease of reference.

\subsection{Preliminaries}
The regional traffic network is represented as a traffic graph $G=(\mathcal{V},\mathcal{A})$, where $\mathcal{V}$ denotes the set of sensor nodes with $N=|\mathcal{V}|$, and $\mathcal{A}\in\mathbb{R}^{N\times N}$ is the spatial adjacency matrix. For each prediction sample, the historical observations and future targets are denoted as $\mathcal{X}\in\mathbb{R}^{\tau_{\mathrm{h}}\times N\times D}$ and $\mathcal{Y}\in\mathbb{R}^{\tau_{\mathrm{f}}\times N\times D}$, where $\tau_{\mathrm{h}}$ and $\tau_{\mathrm{f}}$ are the lengths of the historical and prediction windows, and $D$ is the feature dimension. Each historical tensor is associated with a timestamp $t\in\{1,2,\ldots,T\}$, where $T$ denotes the number of time-of-day indices.

This study investigates cross-domain traffic prediction, which aims to transfer knowledge from a data-rich source region $\mathcal{S}$ to a data-limited target region $\mathcal{T}$. Following the above definitions, we denote the source-domain data as $(G_{\mathcal{S}}, \mathcal{X}_{\mathcal{S}}, \mathcal{Y}_{\mathcal{S}}, t_{\mathcal{S}})$ and the target-domain data as $(G_{\mathcal{T}}, \mathcal{X}_{\mathcal{T}}, \mathcal{Y}_{\mathcal{T}}, t_{\mathcal{T}})$. For each region $r\in\{\mathcal{S},\mathcal{T}\}$, the prediction model takes the historical traffic observations, graph structure, and timestamp as input, and outputs the future traffic states of all sensor nodes:
\begin{equation}
\hat{\mathcal{Y}}_{r}
=
f_{\boldsymbol{\phi}}(\mathcal{X}_{r},G_{r},t_{r})
\end{equation}

At the source-region training stage, the model parameters are optimized on the source domain:
\begin{equation}
\boldsymbol{\phi}_{\mathcal S}
=
\arg\min_{\boldsymbol{\phi}}
\mathbb{E}
\left[
\mathcal{L}_{\mathrm{sc}}
\left(
f_{\boldsymbol{\phi}}(\mathcal{X}_{\mathcal S},G_{\mathcal S},t_{\mathcal S}),
\mathcal{Y}_{\mathcal S}
\right)
\right]
\end{equation}
Then, $\boldsymbol{\phi}_{\mathcal S}$ is used to initialize the target-region model, which is further fine-tuned with the limited target-domain data:
\begin{equation}
\boldsymbol{\phi}_{\mathcal T}
=
\arg\min_{\boldsymbol{\phi};\,\boldsymbol{\phi}^{(0)}=\boldsymbol{\phi}_{\mathcal S}}
\mathbb{E}
\left[
\mathcal{L}_{\mathrm{ft}}
\left(
f_{\boldsymbol{\phi}}(\mathcal{X}_{\mathcal T},G_{\mathcal T},t_{\mathcal T}),
\mathcal{Y}_{\mathcal T}
\right)
\right]
\end{equation}
The final target model $f_{\boldsymbol{\phi}_{\mathcal T}}$ produces multi-step traffic predictions for all sensor nodes in $\mathcal{T}$.

\begin{figure*}[htbp]
    \centering
    \includegraphics[width=0.95\textwidth]{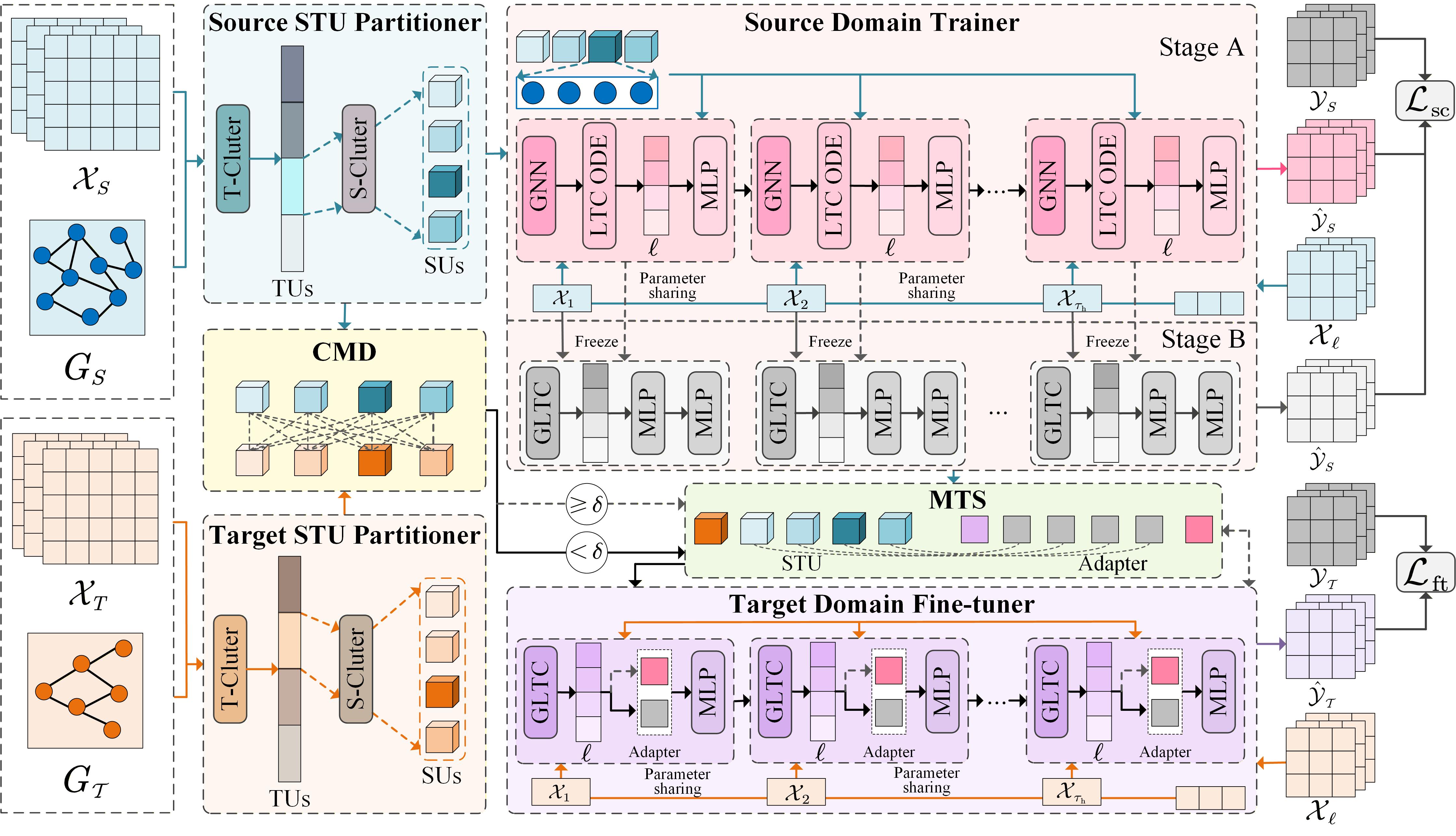}
    \caption{The proposed MA-GLTC architecture.}
    \label{fig:frame}
\end{figure*}

\subsection{MA-GLTC Overview}
As shown in Figure~\ref{fig:frame}, we propose MA-GLTC, a cross-domain traffic prediction framework that enhances spatio-temporal transferability, preserves source-domain knowledge, and captures irregular traffic dynamics. The framework consists of four components. First, the STU partitioning module constructs transferable spatio-temporal units through temporal and spatial clustering. Second, the GLTC-based prediction module models each STU with a GLTC encoder and a time-conditioned decoder, capturing continuous graph-coupled dynamics and generating horizon-aware predictions. Third, the cross-domain learning framework performs source-domain two-stage training, stores transferable STU-level knowledge in MTS, and conducts STU matching and target-domain adaptation. Through this design, MA-GLTC enables fine-grained transfer while preserving source-domain knowledge and adapting to newly emerging target-domain patterns.

\subsection{STU Partitioning Module Design}
The STU partitioning module is designed to characterize the spatio-temporal heterogeneity of traffic data by partitioning the global traffic graph into a set of local STUs. As illustrated in Fig. \ref{fig:frame}, this module consists of two components: temporal clustering (T-Cluster) and spatial clustering (S-Cluster). The details of these two components are presented below.

\subsubsection{Temporal clustering}
T-Cluster is designed to identify functional temporal periods with similar traffic evolution patterns. Given the traffic graph signals organized by time-of-day slots, we first extract statistical features for each time slot to summarize its traffic distribution and fluctuation characteristics. Based on these features, a temporal similarity matrix is constructed, and spectral clustering~\cite{pujulei} is then applied to divide the time-of-day indices into $M$ temporal units: $\{T_m\}_{m=1}^{M}$. Each temporal unit contains time slots with similar traffic dynamics, such as morning peak, evening peak, or off-peak periods. This temporal partition provides a traffic-pattern-aware basis for the following spatial clustering process.

\subsubsection{Spatial clustering}
Given the temporal units $\{T_m\}_{m=1}^{M}$, S-Cluster is performed within each temporal unit to obtain spatial units with similar traffic behaviors. Different from partitioning methods that only rely on road connectivity, S-Cluster mainly considers traffic-pattern similarity among nodes, while incorporating the road topology as a weak structural prior. 

For each temporal unit $T_m$, we extract a traffic-dynamics feature vector $g_i^{(m)}$ for each node $v_i$ to describe its local traffic variation pattern within this period. The traffic-pattern similarity between nodes is denoted as $S^{(m)}_{ij}$. Then, the spatial affinity matrix is constructed as 
\begin{equation} 
W^{\mathrm{spa},m}_{ij} 
= 
(1-\alpha)S^{(m)}_{ij} 
+ 
\alpha \hat{\mathcal{A}}_{ij}
\end{equation} 
where $\hat{\mathcal{A}}$ is the normalized road topology adjacency matrix, and $\alpha\in[0,1]$ controls the contribution of the topology prior. This formulation allows nodes with similar traffic dynamics to be grouped into the same functional unit, while still preserving basic spatial consistency. 

Based on $W^{\mathrm{spa},m}$, Louvain community detection~\cite{Louvain} is applied to obtain the spatial units within each temporal unit: $\{\mathcal{V}_{m,k}\}_{k=1}^{K_m}$. 

Finally, each temporal unit and its corresponding spatial unit are combined to define an STU: 
\begin{equation} 
\mathcal{U}_{m,k}=(T_m,\mathcal{V}_{m,k})
\end{equation} 
The resulting STUs are used as the basic units for subsequent cross-domain alignment and knowledge transfer, each STU is represented as a local graph signal sequence on its corresponding subgraph, enabling fine-grained unit-level modeling.

\subsection{GLTC-based Prediction Module Design}
The GLTC-based prediction module serves as the core predictor of MA-GLTC, consisting of a GLTC encoder and a time-conditioned decoder. The GLTC encoder captures continuous nonlinear spatio-temporal dynamics within each STU, while the time-conditioned decoder incorporates future temporal information to produce horizon-aware residual predictions over the next $\tau_{\mathrm{f}}$ steps.

\subsubsection{GLTC Encoder} 
\begin{figure}[t]
    \centering
    \includegraphics[width=0.42\textwidth]{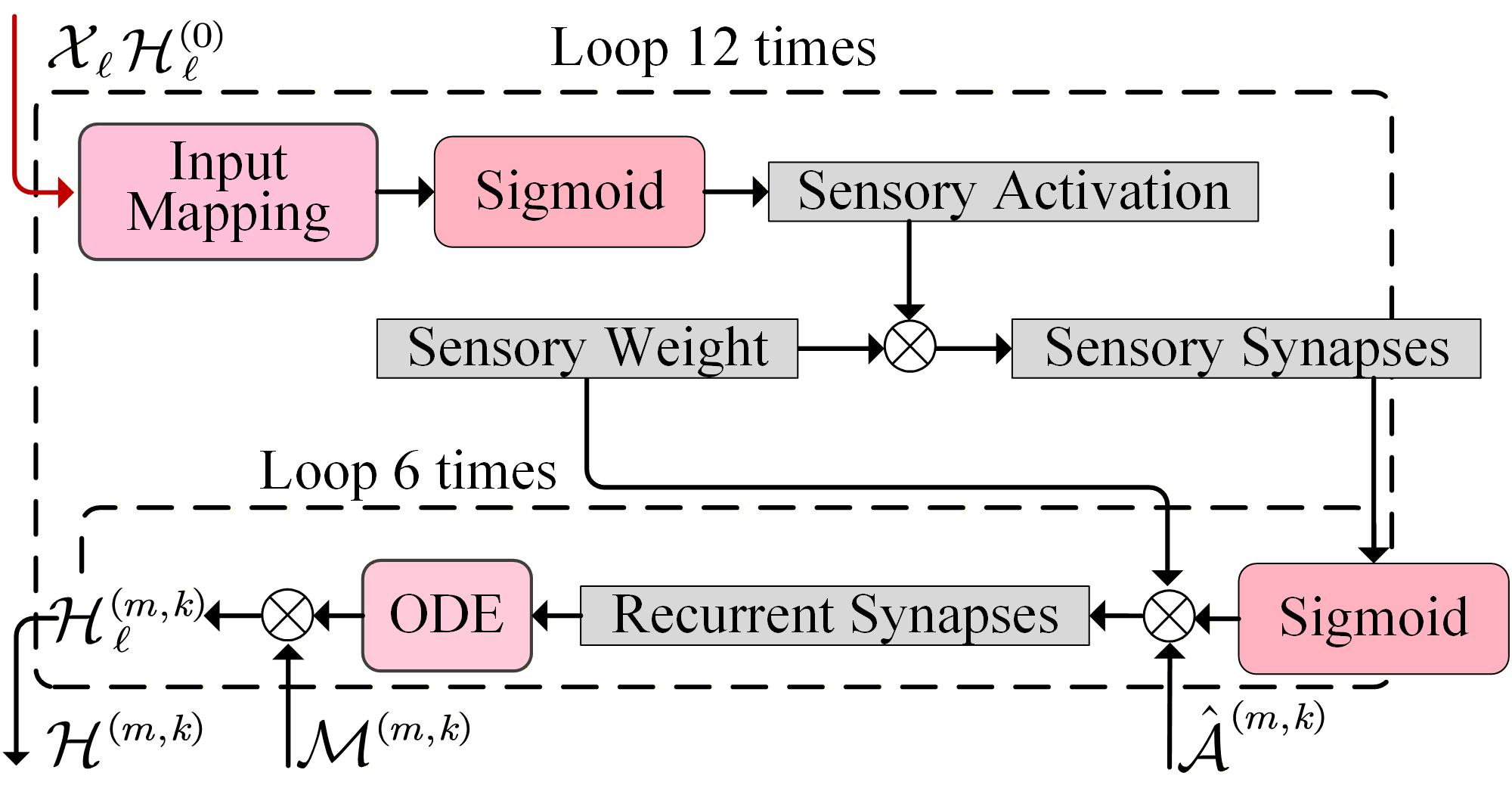}
    \caption{Structure of the GLTC.}
    \label{fig:gltc}
\end{figure}

Classical LTC networks model hidden-state evolution as a continuous-time dynamical system, but they are mainly designed for vector-valued sequences and cannot explicitly capture spatial interactions in graph-structured traffic data. To address this issue, we propose the GLTC encoder, which incorporates graph-coupled recurrent conductance into the LTC dynamics. In this way, the hidden state of each node is updated by its own traffic input, intrinsic temporal memory, and neighborhood feedback from the local STU graph. Fig. \ref{fig:gltc} illustrates the architecture of the GLTC encoder.

For an STU $\mathcal{U}_{m,k}=(T_m,\mathcal{V}_{m,k})$, let $n_{m,k}=|\mathcal{V}_{m,k}|$ denote the number of nodes in the corresponding local subgraph. Given the local historical graph signal $\mathcal{X}^{(m,k)}\in\mathbb{R}^{\tau_{\mathrm{h}}\times n_{m,k}\times D}$, the subgraph adjacency matrix $\mathcal{A}^{(m,k)}\in\mathbb{R}^{n_{m,k}\times n_{m,k}}$, and the node validity mask $\mathcal{M}^{(m,k)}$, GLTC produces the node-level representation: 
\begin{equation}
\mathcal{H}^{(m,k)}
=
\operatorname{GLTC}_{\theta}
\left(
\mathcal{X}^{(m,k)},
\mathcal{A}^{(m,k)},
\mathcal{M}^{(m,k)}
\right)
\label{eq:gltc_encoder_output}
\end{equation}
where $\mathcal{H}^{(m,k)}\in\mathbb{R}^{n_{m,k}\times d_h}$ and $d_h$ is the hidden dimension.

At the $\ell$-th historical step, the hidden state is initialized as $\mathcal{H}_{\ell}^{(0)}=\mathcal{H}_{\ell-1}^{(m,k)}$. To inject spatial interactions into the LTC dynamics, we define the graph-coupled recurrent conductance as
\begin{equation}
g_{\ell,u}^{h}
=
\hat{\mathcal{A}}^{(m,k)}
\sigma_{\mathrm{sig}}
\left(
\left(
\mathcal{H}_{\ell}^{(u)}W_h+b_h-\mu_h
\right)\odot \sigma_h
\right)
\odot w_h
\label{eq:gltc_recurrent_conductance}
\end{equation}
where $\hat{\mathcal{A}}^{(m,k)}$ is the normalized adjacency matrix of the STU-specific subgraph, and $u=0,\ldots,U-1$ denotes the internal unfolding step. Compared with standard LTC, this graph-coupled conductance enables recurrent feedback to be propagated through the local graph structure, allowing each node to receive neighborhood-aware dynamic information during continuous state evolution.

Following the LTC update rule, GLTC integrates the leakage, input-driven, and graph-coupled recurrent terms through a semi-implicit update:
\begin{equation}
\mathcal{H}_{\ell}^{(u+1)}
=
\frac{
\Omega \mathcal{H}_{\ell}^{(u)}
+
g^{l} E^{l}
+
g_{\ell}^{x} E^{x}
+
g_{\ell,u}^{h} E^{h}
}{
\Omega
+
g^{l}
+
g_{\ell}^{x}
+
g_{\ell,u}^{h}
}
\label{eq:gltc_state_update}
\end{equation}
where $\Omega$ denotes the capacitance parameter controlling the inertia of hidden-state evolution, $g_{\ell}^{x}$ denotes the input-driven conductance, $g^{l}$ is the leakage conductance, and $E^{l}$, $E^{x}$, and $E^{h}$ are the corresponding equilibrium potentials. After each update, the mask $\mathcal{M}^{(m,k)}$ is applied to remove the influence of padded virtual nodes. The final hidden state after scanning all $\tau_{\mathrm{h}}$ historical observations is used as the encoded representation of the STU.

The main procedure of the proposed Graph Liquid Time-Constant encoder is summarized in Algorithm~\ref{alg:gltc_encoder}. Given the local historical graph signal and its corresponding subgraph structure, GLTC performs continuous graph-aware state evolution along the temporal dimension. At each observation step, the current traffic input provides input-driven dynamics, while the previous hidden state is propagated through the local graph to generate graph-coupled recurrent feedback. These terms are integrated by a semi-implicit liquid time-constant update, producing the encoded STU representation.

\begin{algorithm}[!t]
\caption{Graph Liquid Time-Constant Encoder}
\label{alg:gltc_encoder}
\small
\begin{algorithmic}[1]
\REQUIRE Historical local graph signal $\mathcal{X}^{(m,k)}$, adjacency matrix $\mathcal{A}^{(m,k)}$, node validity mask $\mathcal{M}^{(m,k)}$, hidden dimension $d_h$
\ENSURE Encoded STU representation $\mathcal{H}^{(m,k)}$

\STATE Initialize hidden state $\mathcal{H}_{0}^{(m,k)}=\mathbf{0}$.
\STATE Normalize the local adjacency matrix to obtain $\hat{\mathcal{A}}^{(m,k)}$.

\FOR{$\ell=1$ to $\tau_{\mathrm{h}}$}
    \STATE Obtain the current traffic observation $\mathcal{X}_{\ell}^{(m,k)}$.
    \STATE Compute the input-driven conductance $g_{\ell}^{x}$ from $\mathcal{X}_{\ell}^{(m,k)}$ using the LTC input pathway.
    \STATE Initialize the internal liquid state as $\mathcal{H}_{\ell}^{(0)}=\mathcal{H}_{\ell-1}^{(m,k)}$.

    \FOR{$u=0$ to $U-1$}
        \STATE Compute the graph-coupled recurrent conductance $g_{\ell,u}^{h}$ using $\mathcal{H}_{\ell}^{(u)}$ and $\hat{\mathcal{A}}^{(m,k)}$ according to Eq.~(\ref{eq:gltc_recurrent_conductance}).
        \STATE Update the liquid state $\mathcal{H}_{\ell}^{(u+1)}$ with $g_{\ell}^{x}$ and $g_{\ell,u}^{h}$ according to Eq.~(\ref{eq:gltc_state_update}).
        \STATE Apply the node validity mask $\mathcal{M}^{(m,k)}$ to suppress padded virtual nodes.
    \ENDFOR

    \STATE Set $\mathcal{H}_{\ell}^{(m,k)}=\mathcal{H}_{\ell}^{(U)}$.
\ENDFOR

\STATE \textbf{return} $\mathcal{H}^{(m,k)}=\mathcal{H}_{\tau_{\mathrm{h}}}^{(m,k)}$.

\end{algorithmic}
\end{algorithm}

\subsubsection{Time-Conditioned Decoder}
Given the encoded STU representation $\mathcal{H}^{(m,k)}$, the time-conditioned decoder generates multi-step predictions by explicitly incorporating future temporal information. Instead of using a fixed output layer for all horizons, the decoder conditions each prediction on the queried future step, allowing the model to capture horizon-dependent traffic variations.

For a future horizon $\Delta$, we construct a future-time encoding $\mathbf{t}_{\Delta}$ using the relative forecast offset and the corresponding time-of-day phase:
\begin{equation}
\mathbf{t}_{\Delta}
=
\left[1,a_{\Delta},a_{\Delta}^{2}\right]
\oplus
\left[
\sin(2\pi r\eta_{\Delta}),
\cos(2\pi r\eta_{\Delta})
\right]_{r=1}^{R}
\end{equation}
where $a_{\Delta}=\Delta/\tau_{\mathrm{f}}$, $\eta_{\Delta}=((t+\Delta-1)\bmod T)/T$, $R$ is the harmonic order, and $\oplus$ denotes feature concatenation. This encoding jointly describes the relative prediction horizon and periodic temporal phase.

To decouple node-specific spatial information from horizon-specific temporal variation, we adopt a factorized decoder. The encoded node representation and future-time encoding are mapped into two latent branches and then combined to generate the residual prediction:
\begin{equation}
\mathcal{Z}^{(m,k)}=
\operatorname{MLP}_{z}
\left(
\mathcal{H}^{(m,k)}
\right)
\end{equation}
\begin{equation}
\mathbf{P}_{\Delta}=
\operatorname{MLP}_{p}
\left(
\mathbf{t}_{\Delta}
\right)
\end{equation}
\begin{equation}
\hat{\mathcal{R}}^{(m,k)}_{:,\Delta}=
\frac{1}{\sqrt{C}}
\mathcal{Z}^{(m,k)}
\mathbf{P}_{\Delta}
+
\mathbf{b}_{\Delta}
\end{equation}
Here, $\mathcal{Z}^{(m,k)}$ encodes node-level traffic characteristics, while $\mathbf{P}_{\Delta}$ and $\mathbf{b}_{\Delta}$ provide future-step-specific temporal modulation.

Considering the strong short-term persistence of traffic states, we introduce a trend baseline to represent stable local variations. Specifically, the baseline is estimated from the most recent $\tau_b$ observations using a clipped linear trend, which avoids unreliable extrapolation caused by abrupt changes. The final prediction is obtained by combining the baseline and residual prediction:
\begin{equation}
\hat{\mathcal{Y}}^{(m,k)}_{:,\Delta}
=
\mathcal{B}^{(m,k)}_{:,\Delta}
+
\hat{\mathcal{R}}^{(m,k)}_{:,\Delta}
\end{equation}
In this formulation, the baseline captures low-frequency short-term trends, while the time-conditioned residual branch focuses on horizon-dependent fluctuations and propagation-related variations.

\subsection{Cross-Domain Learning Framework}
The cross-domain learning framework aims to preserve transferable graph-coupled traffic dynamics learned from the source region while adapting lightweight decoder-side parameters to target-domain heterogeneity. Specifically, the GLTC encoder is used to capture shared continuous traffic dynamics, whereas the time-conditioned decoder is adapted at the STU level to handle local distribution shifts. The framework contains three procedures: source-domain two-stage training, MTS-based memory construction, and target-domain adaptation.

\subsubsection{Source-Domain Two-Stage Training}
In the source domain, we adopt a two-stage strategy to separately learn shared traffic dynamics and STU-specific local characteristics. 
 
\textbf{Stage A: Shared dynamics learning.} A shared GLTC encoder $\operatorname{GLTC}_{\theta}$ and a universal decoder $Dec_{\psi^{\mathrm{uni}}}$ are first trained over all source-domain STUs. For a source STU $\mathcal{U}_{m,k}^{\mathcal{S}}$, the prediction process is written as
\begin{equation}
\begin{cases}
\mathcal{H}_{\mathcal{S}}^{(m,k)}
=
\operatorname{GLTC}_{\theta}
\left(
\mathcal{X}_{\mathcal{S}}^{(m,k)},
\mathcal{A}_{\mathcal{S}}^{(m,k)},
\mathcal{M}_{\mathcal{S}}^{(m,k)}
\right)\\
\hat{\mathcal{Y}}_{\mathcal{S}}^{(m,k)}
=
Dec_{\psi^{\mathrm{uni}}}
\left(
\mathcal{H}_{\mathcal{S}}^{(m,k)},
\mathbf{t}_{\mathcal{S}}^{(m,k)},
\mathcal{B}_{\mathcal{S}}^{(m,k)}
\right)
\end{cases}
\end{equation}

The optimization objective is
\begin{equation}
\{\theta_{\mathcal{S}},\psi_{\mathcal{S}}^{\mathrm{uni}}\}
=
\arg\min_{\theta,\psi^{\mathrm{uni}}}
\mathbb{E}_{\mathcal{U}_{m,k}^{\mathcal{S}}\sim \pi_{\mathcal{S}}}
\left[
\mathcal{L}_{\mathrm{sc}}
\left(
\hat{\mathcal{Y}}_{\mathcal{S}}^{(m,k)},
\mathcal{Y}_{\mathcal{S}}^{(m,k)}
\right)
\right]
\end{equation}
where $\pi_{\mathcal{S}}$ denotes the source STU sampling distribution. To balance data scale and unit coverage, we set $\pi_{\mathcal{S}}(\mathcal{U}_{m,k}^{\mathcal{S}})\propto (n_{m,k}^{\mathcal{S}})^r$, where $n_{m,k}^{\mathcal{S}}$ is the number of samples in $\mathcal{U}_{m,k}^{\mathcal{S}}$ and $0<r<1$.

\textbf{Stage B: STU-specific decoder adaptation.} After Stage A, the GLTC encoder $\theta_{\mathcal{S}}$ is frozen to preserve the learned transferable dynamics. For each source STU, the universal decoder is used to initialize an STU-specific decoder, which is then fine-tuned locally:
\begin{equation}
\psi_\mathcal{S}^{(m,k)}
=\arg\min_{\psi^{(m,k)}}\;
\mathbb{E}
\Big[
\mathcal{L}_\mathrm{sc}\!\big(
\hat{\mathcal{Y}}_{\mathcal{S}}^{(m,k)},\,
\mathcal{Y}_{\mathcal{S}}^{(m,k)}
\big)
\Big]
\end{equation}
This design allows the encoder to retain shared graph-coupled dynamics, while the STU-specific decoders capture local traffic characteristics.

The Huber loss \cite{Huber1992} is used as the basic prediction loss due to its robustness to outliers. Given the prediction error $e=\hat{\mathcal{Y}}-\mathcal{Y}$, the Huber loss is defined as
\begin{equation}
\mathcal{L}_{\mathrm{hub}}(\hat{\mathcal{Y}},\mathcal{Y})
=
\begin{cases}
\frac{1}{2}e^2, & |e|\leq \varepsilon,\\
\varepsilon(|e|-\frac{1}{2}\varepsilon), & |e|>\varepsilon,
\end{cases}
\label{Huber}
\end{equation}
where $\varepsilon$ controls the transition between the quadratic and linear penalty regions. The loss $\mathcal{L}_\mathrm{sc}$ by Eq.(\ref{Huber}) with $\hat{\mathcal{Y}_\mathcal{S}}$ and $\mathcal{Y}_\mathcal{S}$.

\subsubsection{MTS Memory Mechanism}
To preserve transferable source knowledge and support target-domain adaptation, we introduce a MTS mechanism. MTS stores STU-level transfer patterns rather than a single global source model. Each memory entry is defined as
\begin{equation}
\mathbf{MTS}
=
\left\{
(\mathbf{q}_i,\psi_i,\rho_i,n_i,o_i)
\right\}_{i=1}^{L}
\end{equation}
where $\mathbf{q}_i$ is the STU prototype, $\psi_i$ is the corresponding decoder parameter, $\rho_i$ denotes the reliability score, $n_i$ is the number of supporting samples, and $o_i\in\{\mathcal{S},\mathcal{T}\}$ indicates the domain source.

After source-domain training, the universal decoder $\psi_{\mathcal{S}}^{\mathrm{uni}}$ and source STU-specific decoders $\{\psi_{\mathcal{S}}^{(m,k)}\}$ are stored in MTS with their prototypes. For a target STU, its prototype is matched with memory entries using central moment discrepancy (CMD) \cite{cmd}. Matched entries are used to initialize the target decoder with CMD-based weights, while $\psi_{\mathcal{S}}^{\mathrm{uni}}$ is used as a fallback when no reliable match exists.

During target adaptation, MTS is updated only for reliable target patterns. A target entry is merged with an existing target-side entry if their prototypes are close; otherwise, it is added as a new entry. The memory size is bounded by $L_{\max}$, and low-reliability target entries are removed first. Source entries are kept fixed to avoid source-knowledge corruption.

\subsubsection{STU Matching and Target-Domain Adaptation}
Previous methods usually transfer a source-trained model to the target domain as a global initializer~\cite{cmd1,cmd2}, which may ignore the fine-grained correspondence between source and target traffic patterns. To reduce negative transfer, MA-GLTC performs MTS-based STU matching, where each target STU retrieves transferable patterns from the memory entries defined in the previous subsection. CMD is used to measure the distributional discrepancy between the target STU and the stored STU prototypes. The $P$-order CMD is defined as
\begin{equation}
\begin{aligned}
CMD_P(\mathcal{U}^\mathcal{T},\mathcal{U}^\mathcal{S})
=&
\frac{1}{(b-a)}
\left\|
E(\mathcal{U}^\mathcal{T})-E(\mathcal{U}^\mathcal{S})
\right\|_2 \\
&+
\sum_{p=2}^{P}
\frac{1}{(b-a)^p}
\left\|
C_p(\mathcal{U}^\mathcal{T})-C_p(\mathcal{U}^\mathcal{S})
\right\|_2 
\end{aligned}
\end{equation}
where $E(\cdot)$ denotes the expectation, $C_p(\cdot)$ is the $p$-th central moment, and $[a,b]$ is the value range of the samples.

For each target STU, its prototype is compared with the prototypes stored in $\mathcal{M}_{\mathrm{MTS}}$. Memory entries with CMD values lower than the threshold $\delta$ are selected for decoder initialization. When multiple entries are matched, their decoder parameters are combined with CMD-based weights:
\begin{equation}
\psi_{\mathcal{T},0}^{(m,k)}
=
\sum_{CMD_P(\mathbf{q}_{\mathcal{T}}^{(m,k)},\mathbf{q}_i)\leq\delta}
\omega_i\psi_i 
\end{equation}
where $\mathbf{q}_{\mathcal{T}}^{(m,k)}$ is the prototype of the target STU, and $\psi_i$ is the decoder parameter stored in the $i$-th MTS entry. If no reliable memory entry satisfies the threshold, the universal decoder $\psi_{\mathcal{S}}^{\mathrm{uni}}$ is used for initialization. This strategy enables unit-level transfer from MTS and avoids forcing irrelevant source patterns into the target domain.

Following domain adaptation studies~\cite{zuixiaoh}, we introduce covariate alignment to reduce representation-level discrepancy between the target STU and the matched MTS pattern:
\begin{equation}
\mathcal{L}_{\mathrm{cov}}
=
CMD_P
\left(
\mathcal{H}_{\mathcal{T}}^{(m,k)},
\mathbf{q}_i
\right)
\end{equation}
where $\mathcal{H}_{\mathcal{T}}^{(m,k)}$ denotes the target STU representation, and $\mathbf{q}_i$ denotes the matched STU prototype retrieved from MTS.

Since covariate alignment alone cannot ensure label-space consistency, we further introduce label-distribution alignment between the target prediction and observation:
\begin{equation}
\mathcal{L}_{\mathrm{lab}}
=
CMD_P
\left(
\mathcal{Y}_{\mathcal{T}}^{(m,k)},
\hat{\mathcal{Y}}_{\mathcal{T}}^{(m,k)}
\right)
\end{equation}

During target-domain adaptation, the GLTC encoder $\theta_{\mathcal{S}}$ is fixed, and only the target STU-specific decoder is fine-tuned. The optimization objective is formulated as
\begin{equation}
\psi_{\mathcal{T}}^{(m,k)}
=
\arg\min_{\psi^{(m,k)}}
\mathbb{E}
\left[
\mathcal{L}_{\mathrm{ft}}
\left(
\hat{\mathcal{Y}}_{\mathcal{T}}^{(m,k)},
\mathcal{Y}_{\mathcal{T}}^{(m,k)}
\right)
\right]
\end{equation}
where $\psi_{\mathcal{T}}^{(m,k)}$ is initialized by the matched MTS decoder or by $\psi_{\mathcal{S}}^{\mathrm{uni}}$ when no reliable match exists. The fine-tuning loss is defined as
\begin{equation}
\mathcal{L}_{\mathrm{ft}}
=
\mathcal{L}_{\mathrm{hub}}
+
\lambda_1\mathcal{L}_{\mathrm{cov}}
+
\lambda_2\mathcal{L}_{\mathrm{lab}}
\end{equation}
where $\lambda_1$ and $\lambda_2$ control the contributions of covariate alignment and label-distribution alignment, respectively. After fine-tuning, reliable target-domain updates are written back to MTS according to the update rule described above.

\subsection{Computational Complexity Analysis}
The computational cost of MA-GLTC mainly comes from STU partitioning and GLTC-based prediction. For STU partitioning, temporal clustering aggregates traffic features over $N$ nodes and $T$ time slots, with a cost of $\mathcal{O}(NT)$. Spatial clustering computes node-wise similarities and performs community detection, leading to approximately $\mathcal{O}(N^2\log N)$ under the dense similarity setting.

For GLTC-based prediction, the main cost lies in graph-coupled state propagation. Since graph aggregation is performed along the temporal dimension, the prediction complexity is approximately $\mathcal{O}(TN^2)$ under the dense graph setting. The decoder and MTS introduce only lightweight additional costs. Therefore, the overall complexity of MA-GLTC can be written as $\mathcal{O}(NT+N^2\log N+TN^2)$, which is mainly dominated by the GLTC-based graph evolution term $\mathcal{O}(TN^2)$.

\section{EXPERIMENTS AND RESULTS}
This section provides the details of the conducted computational experiments and corresponding results.

\subsection{Experimental Settings}
\subsubsection{Datasets} The MA-GLTC model was evaluated using five open-source traffic datasets, comprising two traffic flow datasets (PeMS04 and PeMS08) \cite{pems} and three traffic speed datasets (PeMS-BAY, DiDi-Chengdu, and DiDi-Shenzhen) \cite{pemsbay,Didi}. Since flow and speed are fundamental metrics characterizing traffic states, incorporating datasets with these distinct features facilitates a robust assessment of the model's generalizability. Key statistics for these five benchmark datasets are summarized in Table \ref{tab:DatasetStats}.

\begin{table}[!t]
\centering
\caption{Dataset statistics}
\label{tab:DatasetStats}
\scriptsize
\resizebox{\columnwidth}{!}{%
\begin{tabular}{c c c c|c c}
\hline\hline
Datasets & PEMS04 & PEMS08 & PeMSBAY & Didi-Chengdu & Didi-Shenzhen \\
\hline
Interval & \multicolumn{3}{c|}{5-min} & \multicolumn{2}{c}{10-min} \\
\hline
samples & 16992 & 17856 & 52116 & 17280 & 17280 \\
nodes   & 307   & 170   & 325   & 524   & 627   \\
edges   & 340   & 295   & 2369  & 1120  & 4845  \\
\hline\hline
\end{tabular}}
\end{table}

\subsubsection{Baselines} We compare MA-GLTC with two groups of baselines: Inner-T and Cross-T.

\begin{itemize}
    \item \textbf{Inner-T baselines.} These methods are trained and evaluated within the target region without explicit cross-domain transfer. HA uses historical averages from corresponding periods for prediction. ARIMA~\cite{arima} models temporal auto-correlation with a statistical time-series formulation. GRU~\cite{gru} captures sequential dependencies through recurrent gating. T-GCN~\cite{tgcn} combines graph convolution with recurrent units for spatio-temporal modeling. DCRNN~\cite{dcrnn} introduces diffusion convolution into an encoder-decoder framework to model traffic propagation. ASTTN~\cite{asttn} learns temporal and dynamic spatial dependencies through attention and adaptive graph convolution. IEEAformer~\cite{ieeaformer} enhances spatio-temporal attention with implicit information embedding and environment-aware temporal modeling.

    \item \textbf{Cross-T baselines.} These methods transfer knowledge from source regions or pre-trained models to improve target-region prediction. DASTNet~\cite{dastnet} adopts domain adversarial learning to reduce source-target distribution discrepancy. STGFSL~\cite{ref2} uses few-shot learning to generate model parameters from traffic meta-knowledge. STGP~\cite{stgp} introduces a task-agnostic prompting framework for multiple traffic prediction tasks. UniST~\cite{unist} builds a unified Transformer-based framework for urban spatio-temporal forecasting. GPD~\cite{gpd} employs diffusion-based generative pre-training to produce target-city model parameters. CGSTT~\cite{CGSTT} performs cluster-level spatio-temporal transfer with dual alignment. MTPB~\cite{MTPB} constructs a multi-scale traffic pattern bank for cross-city knowledge transfer.
\end{itemize}

\subsubsection{Metrics} The performance of the different methods is evaluated using three metrics: mean absolute error (MAE), root mean square error (RMSE), and mean absolute percentage error (MAPE). In the results presented below, the percentage sign ("\%") for MAPE is omitted for the sake of brevity.

\subsubsection{Implementation Details and Experimental Setup} MA-GLTC is implemented in PyTorch and trained with the Adam optimizer on an NVIDIA RTX 3050 Laptop GPU. Source-domain pre-training is conducted for 200 epochs with a learning rate of $1\times10^{-3}$, followed by target-domain fine-tuning for 20 epochs with a learning rate of $1\times10^{-4}$. The batch size, hidden dimension, historical window, and prediction horizon are set to 64, 64, 12, and 12, respectively. The main hyperparameters are set as follows: $\alpha=0.6$, $U=6$, $R=3$, $\beta=0.08$, $\varepsilon=1$, $P=5$, $\delta=0.2$, and $\lambda_1=\lambda_2=1$.

\begin{table*}[!t]
\centering
\caption{Intra-temporal Cross-Domain Traffic Prediction Results}
\label{tab:intra_temporal_results}
\scriptsize 
\renewcommand{\arraystretch}{0.96}
\setlength{\tabcolsep}{12pt}
\begin{tabular}{c|c|c|c c c c c c c c c}
\hline\hline
\multirow{2}{*}{\rotatebox{90}{}} & \multirow{2}{*}{\rotatebox{90}{}} & \multirow{2}{*}{\textbf{Model}} & \multicolumn{3}{c}{\textbf{Horizon 3}} & \multicolumn{3}{c}{\textbf{Horizon 6}} & \multicolumn{3}{c}{\textbf{Horizon 12}} \\
\cline{4-6} \cline{7-9} \cline{10-12}
& & & MAE & RMSE & MAPE & MAE & RMSE & MAPE & MAE & RMSE & MAPE \\
\hline\hline
\multirow{15}{*}{\rotatebox{90}{PeMS04$\rightarrow$PeMS08}} & \multirow{7}{*}{\rotatebox{90}{Inner-T}} & HA & 22.556 & 30.063 & 16.724 & 28.853 & 37.435 & 21.816 & 41.884 & 54.262 & 34.892 \\
 &  & ARIMA & 20.523 & 30.470 & 24.010 & 26.506 & 38.485 & 28.604 & 38.750 & 51.778 & 38.269 \\
 &  & GRU & 19.504 & 28.805 & 13.053 & 21.136 & 31.447 & 15.716 & 27.849 & 38.288 & 21.520 \\
 &  & TGCN & 20.571 & 32.041 & 12.125 & 26.519 & 40.429 & 16.433 & 37.396 & 54.148 & 24.522 \\
 &  & DCRNN & 17.758 & 27.979 & 12.281 & 19.485 & 30.486 & 13.774 & 26.046 & 39.907 & 18.170 \\
 &  & ASTTN & 15.845 & 25.446 & 10.279 & 17.016 & 27.311 & 11.033 & 22.827 & 35.586 & 15.646 \\
 &  & IEEAFormer & 15.551 & 25.088 & 10.208 & \underline{16.868} & 27.205 & 11.161 & 21.437 & 33.508 & 16.390 \\
 \cline{2-12}
 & \multirow{8}{*}{\rotatebox{90}{Cross-T}} & DASTNet & 17.139 & 26.612 & 11.965 & 19.549 & 30.335 & 14.698 & 24.267 & 36.620 & 17.177 \\
 &  & STGFSL & 18.254 & 28.889 & 15.540 & 21.880 & 34.084 & 18.955 & 29.967 & 44.473 & 25.548 \\
 &  & STGP & 17.346 & 24.979 & 15.707 & 20.256 & 28.935 & 16.552 & 28.563 & 38.721 & 23.081 \\
 &  & UniST & 21.168 & 36.511 & 13.704 & 24.979 & 40.602 & 16.044 & 30.356 & 44.704 & 25.416 \\
 &  & GDP & 21.102 & 26.383 & \textbf{9.726} & 23.648 & 29.903 & \textbf{10.331} & 29.150 & 36.273 & \underline{13.174} \\
 &  & CGSTT & \underline{15.461} & \underline{24.091} & 10.623 & 17.098 & \underline{26.754} & 12.163 & \underline{20.599} & \underline{31.879} & 14.641 \\
 &  & MTPB & 18.165 & 28.537 & 16.640 & 21.296 & 33.278 & 20.070 & 27.866 & 42.030 & 28.572 \\
 &  & MA-GLTC & \textbf{15.124} & \textbf{23.325} & \underline{9.820} & \textbf{16.359} & \textbf{25.503} & \underline{10.588} & \textbf{18.927} & \textbf{29.478} & \textbf{12.445} \\
\hline\hline
\multirow{15}{*}{\rotatebox{90}{PeMS08$\rightarrow$PeMS04}} & \multirow{7}{*}{\rotatebox{90}{Inner-T}} & HA & 26.040 & 40.274 & 20.760 & 33.918 & 48.313 & 25.036 & 47.551 & 63.078 & 37.871 \\
 &  & ARIMA & 23.774 & 36.927 & 12.321 & 31.210 & 48.706 & 15.327 & 42.594 & 63.725 & 19.700 \\
 &  & GRU & 23.582 & 30.914 & 13.876 & 26.705 & 35.367 & 17.451 & 32.510 & 47.013 & 26.613 \\
 &  & TGCN & 23.260 & 32.468 & 11.268 & 29.112 & 41.312 & 12.986 & 41.662 & 57.250 & 15.343 \\
 &  & DCRNN & 21.429 & 30.847 & 13.339 & 24.873 & \underline{33.032} & 14.448 & 31.539 & 43.668 & 20.582 \\
 &  & ASTTN & 20.327 & 31.972 & 13.693 & 21.727 & 34.084 & 14.716 & 26.919 & \textbf{37.247} & 17.978 \\
 &  & IEEAFormer & 20.068 & 32.062 & 13.242 & 21.437 & 33.508 & 16.395 & 26.046 & 39.907 & 18.171 \\
 \cline{2-12}
 & \multirow{8}{*}{\rotatebox{90}{Cross-T}} & DASTNet & 20.837 & 32.632 & 16.368 & 23.834 & 37.006 & 19.311 & 30.237 & 45.656 & 24.086 \\
 &  & STGFSL & 21.718 & 33.403 & 17.512 & 25.018 & 38.304 & 21.074 & 32.514 & 48.643 & 29.987 \\
 &  & STGP & 21.238 & \underline{30.144} & \underline{10.436} & 24.345 & 34.312 & \underline{12.154} & 31.528 & 43.495 & \underline{15.331} \\
 &  & UniST & 24.963 & 40.586 & 16.070 & 29.226 & 45.121 & 20.228 & 38.829 & 53.683 & 28.772 \\
 &  & GDP & 23.845 & 35.804 & 15.753 & 28.592 & 41.937 & 16.857 & 37.898 & 53.833 & 18.530 \\
 &  & CGSTT & \textbf{19.546} & 30.921 & 13.045 & \underline{21.304} & 33.447 & 14.083 & \underline{24.834} & 38.182 & 16.666 \\
 &  & MTPB & 23.685 & 37.320 & 20.749 & 27.718 & 42.769 & 24.204 & 36.414 & 54.461 & 34.280 \\
 &  & MA-GLTC & \underline{19.698} & \textbf{28.632} & \textbf{10.223} & \textbf{21.234} & \textbf{32.952} & \textbf{11.267} & \textbf{24.708} & \underline{37.630} & \textbf{15.255} \\
\hline\hline
\multirow{15}{*}{\rotatebox{90}{Chengdu$\rightarrow$Shenzhen}} & \multirow{7}{*}{\rotatebox{90}{Inner-T}} & HA & 3.127 & 4.043 & 18.338 & 3.414 & 4.476 & 20.686 & 3.946 & 5.177 & 23.924 \\
 &  & ARIMA & 2.631 & 3.214 & 7.753 & 3.037 & 3.767 & \textbf{8.951} & 3.528 & 4.573 & 10.980 \\
 &  & GRU & 2.376 & 3.528 & 9.474 & 2.707 & 4.007 & 10.794 & 3.126 & 4.561 & 12.412 \\
 &  & TGCN & 2.996 & 4.181 & 12.430 & 3.209 & 4.560 & 13.375 & 3.583 & 5.161 & 14.995 \\
 &  & DCRNN & 2.236 & 3.426 & 9.497 & 2.536 & 3.962 & 11.012 & 2.908 & 4.567 & 12.699 \\
 &  & ASTTN & 2.193 & 3.381 & 9.342 & 2.361 & 3.675 & 10.087 & 2.594 & 4.044 & 11.119 \\
 &  & IEEAFormer & 1.992 & 3.121 & 8.465 & \underline{2.196} & \underline{3.371} & 8.989 & 2.484 & \underline{3.542} & \textbf{9.424} \\
 \cline{2-12}
 & \multirow{8}{*}{\rotatebox{90}{Cross-T}} & DASTNet & 2.048 & 7.318 & 12.312 & 2.431 & 8.143 & 15.927 & 3.597 & 11.826 & 23.327 \\
 &  & STGFSL & 2.470 & 3.671 & 10.283 & 2.893 & 4.314 & 12.044 & 3.475 & 5.119 & 14.379 \\
 &  & STGP & 2.469 & 3.683 & 9.875 & 2.814 & 4.212 & 11.469 & 3.309 & 4.884 & 13.662 \\
 &  & UniST & 2.655 & 3.745 & 11.588 & 2.879 & 4.516 & 12.594 & 3.666 & 5.665 & 15.938 \\
 &  & GDP & 2.562 & 5.793 & 26.144 & 2.804 & 6.002 & 27.530 & 3.513 & 6.401 & 30.039 \\
 &  & CGSTT & 2.085 & 3.250 & 8.911 & 2.229 & 3.554 & 9.634 & \textbf{2.395} & 3.867 & \underline{10.448} \\
 &  & MTPB & \underline{1.935} & \underline{2.868} & \underline{7.567} & 2.383 & 3.574 & 9.216 & 2.971 & 4.430 & 11.680 \\
 &  & MA-GLTC & \textbf{1.841} & \textbf{2.613} & \textbf{7.026} & \textbf{2.102} & \textbf{2.993} & \underline{8.980} & \underline{2.435} & \textbf{3.462} & 10.620 \\
\hline\hline
\multirow{15}{*}{\rotatebox{90}{Shenzhen$\rightarrow$Chengdu}} & \multirow{7}{*}{\rotatebox{90}{Inner-T}} & HA & 3.276 & 5.920 & 17.951 & 3.667 & 6.310 & 18.689 & 4.559 & 6.869 & 19.817 \\
 &  & ARIMA & 2.970 & 3.775 & 10.734 & 3.334 & 4.257 & 11.679 & 3.954 & 5.138 & 13.865 \\
 &  & GRU & 2.705 & 3.967 & 11.732 & 3.030 & 4.424 & 13.431 & 3.498 & 5.011 & 15.699 \\
 &  & TGCN & 2.737 & 3.936 & 12.234 & 3.019 & 4.342 & 13.672 & 3.489 & 4.983 & 15.969 \\
 &  & DCRNN & 2.646 & 3.904 & 11.981 & 3.083 & 4.547 & 14.229 & 3.775 & 5.454 & 17.561 \\
 &  & ASTTN & 2.474 & 3.693 & 11.248 & 2.662 & 3.969 & 12.222 & 2.938 & 4.329 & 13.598 \\
 &  & IEEAFormer & \underline{2.223} & \underline{3.406} & \underline{10.120} & \underline{2.381} & 3.932 & \underline{11.620} & 2.913 & 4.358 & \underline{12.355} \\
 \cline{2-12}
 & \multirow{8}{*}{\rotatebox{90}{Cross-T}} & DASTNet & 2.258 & 3.472 & 11.379 & 2.846 & 4.221 & 13.040 & 3.961 & 5.105 & 18.844 \\
 &  & STGFSL & 2.783 & 4.031 & 12.346 & 3.263 & 4.703 & 14.682 & 4.029 & 5.680 & 18.352 \\
 &  & STGP & 2.560 & 3.861 & 13.898 & 3.184 & 4.628 & 15.900 & 3.965 & 5.401 & 17.902 \\
 &  & UniST & 2.651 & 3.951 & 10.665 & 2.978 & 4.326 & 11.854 & 3.839 & 9.168 & 26.806 \\
 &  & GDP & 2.805 & 3.792 & 21.297 & 3.195 & 4.319 & 24.257 & 3.789 & 5.074 & 28.808 \\
 &  & CGSTT & 2.374 & 3.578 & 10.981 & 2.553 & \underline{3.868} & 12.085 & \underline{2.847} & \underline{4.317} & 13.831 \\
 &  & MTPB & 2.375 & 3.422 & \textbf{9.836} & 2.790 & 4.378 & 13.087 & 3.702 & 5.288 & 16.286 \\
 &  & MA-GLTC & \textbf{2.086} & \textbf{3.216} & 10.868 & \textbf{2.322} & \textbf{3.593} & \textbf{11.450} & \textbf{2.845} & \textbf{4.074} & \textbf{12.047} \\
\hline\hline
\end{tabular}
\begin{flushleft}
\vspace{-2pt}
\textit{Note: The best performance is highlighted in \textbf{bold}, and the second-best results are indicated by \underline{underlining}. \\
The percentage symbol (\%) for MAPE is omitted for brevity.}
\end{flushleft}
\end{table*}
\begin{table*}[t]
\centering
\caption{Inter-temporal Cross-Domain Traffic Prediction Results}
\label{tab:inter_temporal_results}
\scriptsize 
\renewcommand{\arraystretch}{0.96}
\setlength{\tabcolsep}{12pt}
\begin{tabular}{c|c|c|c c c c c c c c c}
\hline\hline
\multirow{2}{*}{\rotatebox{90}{}} & \multirow{2}{*}{\rotatebox{90}{}} & \multirow{2}{*}{\textbf{Model}} & \multicolumn{3}{c}{\textbf{Horizon 3}} & \multicolumn{3}{c}{\textbf{Horizon 6}} & \multicolumn{3}{c}{\textbf{Horizon 12}} \\
\cline{4-6} \cline{7-9} \cline{10-12}
& & & MAE & RMSE & MAPE & MAE & RMSE & MAPE & MAE & RMSE & MAPE \\
\hline\hline
\multirow{15}{*}{\rotatebox{90}{Chengdu$\rightarrow$PeMSBAY}} & \multirow{7}{*}{\rotatebox{90}{Inner-T}} & HA & 2.804 & 5.917 & 12.927 & 3.275 & 6.146 & 16.967 & 3.857 & 8.532 & 22.639 \\
 &  & ARIMA & 2.250 & 4.595 & 9.905 & 2.529 & 5.893 & 11.222 & 3.470 & 7.505 & 15.385 \\
 &  & GRU & 2.347 & 3.869 & 13.069 & 2.689 & \underline{4.271} & 17.577 & 2.929 & 5.460 & 23.402 \\
 &  & TGCN & 2.566 & 4.277 & 5.455 & 2.807 & 4.851 & 6.221 & 3.172 & 5.561 & 7.337 \\
 &  & DCRNN & 2.394 & 4.947 & 5.953 & 2.802 & 5.081 & 7.146 & 3.272 & 6.171 & 8.577 \\
 &  & ASTTN & 1.851 & 4.141 & 4.338 & 2.287 & 5.132 & 6.363 & 2.934 & 5.904 & 7.579 \\
 &  & IEEAFormer & \underline{1.683} & 3.818 & 3.980 & \underline{1.986} & 4.736 & 5.700 & \underline{2.383} & 5.723 & 7.509 \\
 \cline{2-12}
 & \multirow{8}{*}{\rotatebox{90}{Cross-T}} & DASTNet & 1.943 & 3.585 & 7.584 & 2.208 & 4.857 & 10.875 & 3.218 & 5.461 & 13.207 \\
 &  & STGFSL & 1.791 & 3.423 & 14.377 & 2.270 & 4.607 & 15.822 & 2.959 & 6.046 & 16.118 \\
 &  & STGP & 1.966 & 4.764 & 17.779 & 2.347 & 5.190 & 19.032 & 3.536 & 7.166 & 23.168 \\
 &  & UniST & 2.621 & 3.709 & 11.495 & 2.822 & 4.272 & 12.347 & 3.340 & \underline{5.427} & 15.397 \\
 &  & GDP & 2.354 & 5.781 & 15.493 & 2.572 & 6.130 & 19.821 & 2.950 & 6.632 & 24.143 \\
 &  & CGSTT & 1.774 & 3.591 & 3.793 & 2.268 & 4.795 & \underline{5.643} & 3.026 & 6.314 & \underline{7.050} \\
 &  & MTPB & 1.688 & \textbf{3.253} & \textbf{3.481} & 2.275 & 4.647 & 5.733 & 3.121 & 6.455 & 7.608 \\
 &  & MA-GLTC & \textbf{1.457} & \underline{3.258} & \underline{3.593} & \textbf{1.717} & \textbf{4.178} & \textbf{5.328} & \textbf{2.365} & \textbf{5.416} & \textbf{6.680} \\
\hline\hline
\multirow{15}{*}{\rotatebox{90}{Shenzhen$\rightarrow$PeMSBAY}} & \multirow{7}{*}{\rotatebox{90}{Inner-T}} & HA & 2.804 & 5.917 & 12.927 & 3.275 & 6.146 & 16.967 & 3.857 & 8.532 & 22.639 \\
 &  & ARIMA & 2.250 & 4.595 & 9.905 & 2.529 & 5.893 & 11.222 & 3.470 & 7.505 & 15.385 \\
 &  & GRU & 2.347 & 3.869 & 13.069 & 2.689 & \underline{4.271} & 17.577 & 2.929 & \underline{5.460} & 23.402 \\
 &  & TGCN & 2.566 & 4.277 & 5.455 & 2.807 & 4.851 & 6.221 & 3.172 & 5.561 & 7.337 \\
 &  & DCRNN & 2.394 & 4.947 & 5.953 & 2.802 & 5.081 & 7.146 & 3.272 & 6.171 & 8.577 \\
 &  & ASTTN & 1.851 & 4.141 & 4.338 & 2.287 & 5.132 & 6.363 & 2.934 & 5.904 & 7.579 \\
 &  & IEEAFormer & 1.683 & 3.818 & 3.980 & \underline{1.986} & 4.736 & 5.700 & \textbf{2.383} & 5.723 & 7.509 \\
 \cline{2-12}
 & \multirow{8}{*}{\rotatebox{90}{Cross-T}} & DASTNet & 2.256 & 4.286 & 12.493 & 2.515 & 5.386 & 15.889 & 3.219 & 5.654 & 16.624 \\
 &  & STGFSL & 1.664 & 3.823 & 14.325 & 2.133 & 4.718 & 15.959 & 2.825 & 6.321 & 16.930 \\
 &  & STGP & \underline{1.624} & 4.325 & 18.530 & 2.071 & 4.537 & 19.322 & 2.820 & 7.514 & 22.304 \\
 &  & UniST & 2.520 & 4.337 & 11.908 & 2.861 & 5.886 & 12.392 & 3.598 & 6.461 & 14.234 \\
 &  & GDP & 2.581 & 4.157 & 18.112 & 2.990 & 4.835 & 20.922 & 3.583 & 5.794 & 25.093 \\
 &  & CGSTT & 1.665 & \underline{3.422} & \underline{3.672} & 2.158 & 4.633 & \underline{5.043} & 2.874 & 6.119 & \underline{7.006} \\
 &  & MTPB & 1.757 & \textbf{3.275} & 3.698 & 2.387 & 4.718 & 5.461 & 3.228 & 6.539 & 7.525 \\
 &  & MA-GLTC & \textbf{1.326} & 3.654 & \textbf{3.596} & \textbf{1.635} & \textbf{4.133} & \textbf{4.475} & \underline{2.404} & \textbf{5.395} & \textbf{6.524} \\
\hline\hline
\multirow{15}{*}{\rotatebox{90}{PeMSBAY$\rightarrow$Chengdu}} & \multirow{7}{*}{\rotatebox{90}{Inner-T}} & HA & 3.276 & 5.920 & 17.951 & 3.667 & 6.310 & 18.689 & 4.559 & 6.869 & 19.817 \\
 &  & ARIMA & 2.970 & 3.775 & 10.734 & 3.334 & 4.257 & 11.679 & 3.954 & 5.138 & 13.865 \\
 &  & GRU & 2.705 & 3.967 & 11.732 & 3.030 & 4.424 & 13.431 & 3.498 & 5.011 & 15.699 \\
 &  & TGCN & 2.737 & 3.936 & 12.234 & 3.019 & 4.342 & 13.672 & 3.489 & 4.983 & 15.969 \\
 &  & DCRNN & 2.646 & 3.904 & 11.981 & 3.083 & 4.547 & 14.229 & 3.775 & 5.454 & 17.561 \\
 &  & ASTTN & 2.474 & 3.693 & 11.248 & 2.662 & 3.969 & 12.222 & \underline{2.938} & \underline{4.329} & 13.598 \\
 &  & IEEAFormer & 2.223 & 3.406 & 10.120 & \textbf{2.381} & 3.932 & \underline{11.620} & \textbf{2.913} & 4.358 & \textbf{12.355} \\
 \cline{2-12}
 & \multirow{8}{*}{\rotatebox{90}{Cross-T}} & DASTNet & \underline{2.162} & \underline{3.287} & 12.785 & 2.607 & \textbf{3.383} & 13.550 & 3.530 & 5.159 & 16.785 \\
 &  & STGFSL & 2.920 & 4.215 & 13.625 & 3.322 & 4.795 & 15.634 & 4.008 & 5.677 & 18.887 \\
 &  & STGP & 2.826 & 3.781 & 13.423 & 3.230 & 4.803 & 14.300 & 3.967 & 5.490 & 18.343 \\
 &  & UniST & 2.711 & 4.901 & 16.406 & 3.153 & 5.289 & 17.360 & 4.028 & 7.367 & 20.898 \\
 &  & GDP & 2.826 & 6.673 & 18.233 & 3.214 & 7.122 & 18.988 & 3.859 & 7.847 & 21.092 \\
 &  & CGSTT & 2.573 & 3.836 & 12.012 & 2.816 & 4.217 & 13.341 & 3.194 & 4.749 & 15.392 \\
 &  & MTPB & 2.360 & 3.415 & \underline{9.568} & 2.820 & 4.336 & 12.808 & 3.668 & 5.219 & 16.110 \\
 &  & MA-GLTC & \textbf{2.148} & \textbf{3.279} & \textbf{9.444} & \underline{2.386} & \underline{3.660} & \textbf{10.202} & 2.942 & \textbf{4.320} & \underline{12.524} \\
\hline\hline
\multirow{15}{*}{\rotatebox{90}{PeMSBAY$\rightarrow$Shenzhen}} & \multirow{7}{*}{\rotatebox{90}{Inner-T}} & HA & 3.127 & 4.043 & 18.338 & 3.414 & 4.476 & 20.686 & 3.946 & 5.177 & 23.924 \\
 &  & ARIMA & 2.631 & 3.214 & 7.753 & 3.037 & 3.767 & \underline{8.951} & 3.528 & 4.573 & 10.980 \\
 &  & GRU & 2.376 & 3.528 & 9.474 & 2.707 & 4.007 & 10.794 & 3.126 & 4.561 & 12.412 \\
 &  & TGCN & 2.996 & 4.181 & 12.430 & 3.209 & 4.560 & 13.375 & 3.583 & 5.161 & 14.995 \\
 &  & DCRNN & 2.236 & 3.426 & 9.497 & 2.536 & 3.962 & 11.012 & 2.908 & 4.567 & 12.699 \\
 &  & ASTTN & 2.193 & 3.381 & 9.342 & 2.361 & 3.675 & 10.087 & 2.594 & 4.044 & 11.119 \\
 &  & IEEAFormer & 1.992 & 3.121 & 8.465 & \underline{2.196} & \underline{3.371} & 8.989 & \underline{2.484} & \textbf{3.542} & \textbf{9.424} \\
 \cline{2-12}
 & \multirow{8}{*}{\rotatebox{90}{Cross-T}} & DASTNet & 2.318 & 3.548 & 8.162 & 2.660 & 4.465 & 11.138 & 3.344 & 5.796 & 15.172 \\
 &  & STGFSL & 2.526 & 3.668 & 10.212 & 2.950 & 4.272 & 11.952 & 3.404 & 4.947 & 14.049 \\
 &  & STGP & 2.171 & 3.366 & 9.735 & 2.569 & 4.138 & 11.870 & 3.304 & 4.735 & 13.410 \\
 &  & UniST & 2.734 & 4.562 & 12.906 & 3.135 & 5.901 & 13.367 & 3.917 & 6.901 & 16.406 \\
 &  & GDP & 2.588 & 6.449 & 17.574 & 3.026 & 6.920 & 18.371 & 3.701 & 7.698 & 19.597 \\
 &  & CGSTT & 2.312 & 3.764 & 11.906 & 2.644 & 4.422 & 13.314 & 3.148 & 4.962 & 15.376 \\
 &  & MTPB & \textbf{1.866} & \underline{2.882} & \underline{7.583} & 2.354 & 3.505 & 9.146 & 3.488 & 4.288 & 11.435 \\
 &  & MA-GLTC & \underline{1.898} & \textbf{2.777} & \textbf{6.991} & \textbf{2.098} & \textbf{3.214} & \textbf{7.564} & \textbf{2.459} & \underline{3.754} & \underline{9.931} \\
\hline\hline
\end{tabular}
\end{table*}

\subsection{Results and analysis}
This study predicts traffic conditions for the next 12 time steps. Tables \ref{tab:intra_temporal_results} and \ref{tab:inter_temporal_results} present the cross-domain forecasting results using each of the five datasets as the source domain. The best performance are highlighted in bold, and the underlined data indicate sub-optimal results.

\subsubsection{Same-Step Flow Prediction}
Table~\ref{tab:intra_temporal_results} reports the intra-temporal cross-domain prediction results for traffic flow and speed. For the flow prediction tasks, i.e., PeMS04$\to$PeMS08 and PeMS08$\to$PeMS04, MA-GLTC consistently achieves the best performance across both short-term and long-term horizons. This indicates that the proposed model can effectively transfer traffic knowledge between different cities and maintain stable prediction accuracy under varying forecasting lengths. Among the Inner-T baselines, statistical and recurrent methods such as HA, ARIMA, and GRU mainly capture temporal regularities and lack sufficient spatial modeling ability. Graph-based spatio-temporal models, including T-GCN and DCRNN, introduce road-network dependencies but are still limited by fixed graph propagation or recurrent temporal modeling. Attention-based methods such as ASTTN and IEEAformer improve dynamic dependency learning, yet they are trained within the target region and do not explicitly address cross-domain distribution shifts. For Cross-T baselines, DASTNet and ST-GFSL show limited gains due to insufficient spatio-temporal representation and alignment capability. STGP, UniST, and GPD improve transferability through prompting, unified modeling, or generative pre-training, but they still lack fine-grained adaptation to local spatio-temporal heterogeneity. Recent transfer models such as CGSTT and MTPB further exploit cluster-level or pattern-bank knowledge, yet their adaptation to unseen target-domain patterns remains limited. Compared with these baselines, MA-GLTC achieves average improvements of 2.23\%, 2.81\%, and 1.99\% in MAE, RMSE, and MAPE, respectively. The superior performance can be attributed to three aspects: STU partitioning enables fine-grained unit-level transfer, GLTC captures continuous graph-coupled traffic dynamics, and MTS preserves source-domain knowledge while adapting to emerging target-domain patterns.

\begin{figure*}[t]
    \centering
    \includegraphics[width=0.9\textwidth]{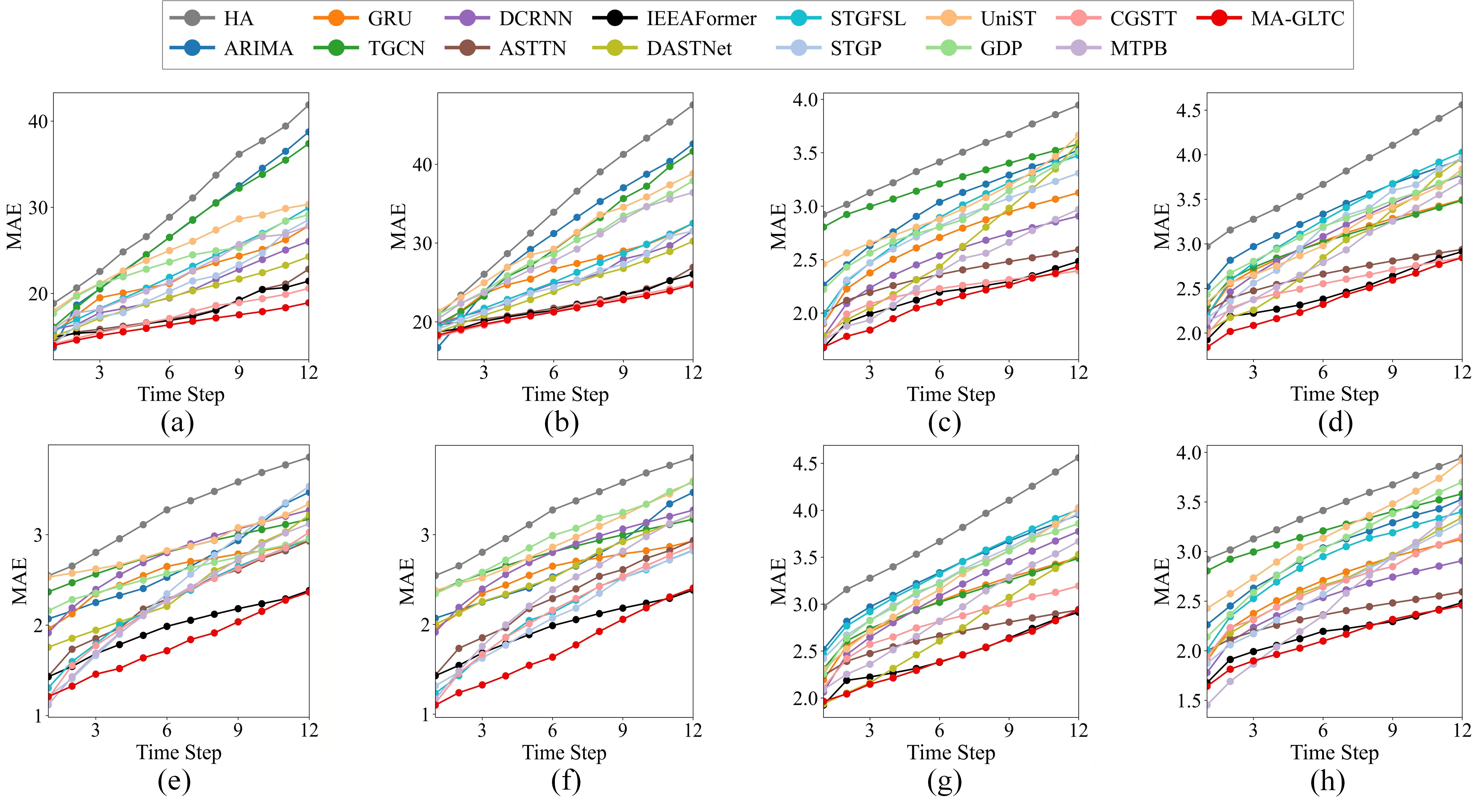}
    \caption{Twelve-time-step prediction results of the baselines and MA-GLTC. (a) PeMS04$\to$PeMS08. (b) PeMS08$\to$PeMS04. (c) Chengdu$\to$Shenzhen. (d) Shenzhen$\to$Chengdu. (e) Chengdu$\to$PeMSBAY. (f) Shenzhen$\to$PeMSBAY. (g) PeMSBAY$\to$Chengdu. (h) PeMSBAY$\to$Shenzhen.}
    \label{fig:qushi}
\end{figure*}

\subsubsection{Same-Step Speed Prediction}
Compared with traffic flow, traffic speed is more strongly constrained by road speed limits and is sensitive to abrupt transitions between free-flow and congested states. As shown in Table~\ref{tab:intra_temporal_results}, MA-GLTC consistently achieves superior performance on the Chengdu$\to$Shenzhen and Shenzhen$\to$Chengdu speed prediction tasks. Traditional spatio-temporal models, such as DCRNN and ASTTN, can capture general temporal trends but tend to smooth rapid speed variations, leading to limited performance under sudden congestion changes. Cross-domain baselines improve transferability to some extent, but they may still overlook target-city-specific congestion thresholds and local speed fluctuations. In contrast, MA-GLTC achieves average MAE improvements of 2.49\% and 2.92\% on the two transfer directions, respectively. This improvement is mainly attributed to the continuous dynamic modeling ability of GLTC and the MTS mechanism, which helps retain transferable source patterns while adapting to target-specific congestion behaviors.

\subsubsection{Cross-Step Speed Prediction}
Cross-step prediction poses a more challenging setting for cross-domain traffic prediction, since source and target domains may have different sampling frequencies and prediction horizons. As shown in Table~\ref{tab:inter_temporal_results}, most baselines exhibit clear performance degradation under temporal granularity mismatch. Compared with same-step transfer, the average MAE of eight cross-domain baselines increases by 20.10\% in the Chengdu$\to$PeMSBAY task and by 6.64\% in the Shenzhen$\to$PeMSBAY task. These results indicate that cross-step prediction introduces additional temporal alignment difficulty beyond spatial domain discrepancy. Existing cross-domain methods are often built on fixed discrete input lengths or predefined temporal structures. Therefore, when facing heterogeneous time resolutions, they usually require resampling, aggregation, or padding to align the input sequences. Such hard temporal alignment may weaken high-frequency traffic variations and distort the original evolution patterns. Although recent methods such as UniST, GPD, CGSTT, and MTPB improve cross-domain transferability, they still show limited flexibility when temporal granularity mismatch and unseen target-domain patterns appear simultaneously.

In contrast, MA-GLTC maintains stable performance in cross-step scenarios. The GLTC module models traffic evolution in a continuous-time manner, reducing its dependence on fixed discrete intervals. These analyses demonstrate the robustness and generalizability of MA-GLTC across diverse traffic datasets, validating its capability to capture complex spatial dependencies and continuous-time temporal dynamics. As shown in Fig. \ref{fig:qushi}, prediction errors generally increase with longer horizons, but MA-GLTC shows smaller performance degradation in most cases, demonstrating its robustness in long-term cross-step prediction.

\subsubsection{Unseen Patterns}
Source and target datasets may contain heterogeneous spatio-temporal patterns due to differences in road structures, traffic demand, and sampling environments. When these discrepancies are large, unseen traffic patterns may emerge in the target domain and weaken cross-domain transfer. MA-GLTC handles unseen patterns through MTS. If no matched STU pattern is retrieved, the universal decoder is used for initialization, and reliable target-domain updates are written back to MTS after fine-tuning. Table~\ref{tab:Unseen Patterns} reports the number of unseen patterns, measured by the increase in STUs. Transfer pairs with similar distributions generate fewer unseen patterns, while heterogeneous pairs require more additional STUs.

\begin{table}[t]
\centering
\caption{Number of Unseen Patterns Identified in Cross-Domain Experiments}
\label{tab:Unseen Patterns}
\setlength{\tabcolsep}{5pt}
\begin{tabular}{c|cc|ccc}
\hline\hline
    Source & PeMS04 & PeMS08 & Chengdu & Shenzhen & PeMSBAY \\
    \hline
    PeMS04   & 0 & 0 &   &   &   \\
    PeMS08   & 2 & 0 &   &   &   \\
    \cline{2-6}
    Chengdu  &   &   & 0 & 0 & 5 \\
    Shenzhen &   &   & 1 & 0 & 5 \\
    PeMSBAY  &   &   & 36 & 8 & 0 \\
    \hline\hline
    \end{tabular}
\end{table}

\subsubsection{Ablation study}
To evaluate the contribution of each component in MA-GLTC, we conduct ablation studies with the following variants:

\begin{itemize}
    \item w/o TU: Removes temporal unit partitioning.
    \item w/o SU: Removes spatial unit partitioning.
    \item w/o STU: Removes the entire Spatio-Temporal Unit partitioning module.
    \item w/o Semi-implicit (Explicit Euler/RK4): Replaces the default semi-implicit numerical solver for continuous dynamics with either the lower-complexity Explicit Euler method or the higher-order Fourth-order Runge-Kutta (RK4) method.
    \item w/o MTS: removes the MTS mechanism for unseen pattern adaptation.
\end{itemize}

\begin{table*}[t]
\centering
\caption{Ablation Study of MA-GLTC and Its Variants at Horizon 12}
\label{tab:ablation}
\scriptsize 
\renewcommand{\arraystretch}{1}
\setlength{\tabcolsep}{7pt} 
\begin{tabular}{l|c c c|c c c|c c c|c c c}
\hline\hline
\multirow{2}{*}{\textbf{Variant}} & \multicolumn{3}{c|}{\textbf{PeMS04$\to$PeMS08}} & \multicolumn{3}{c|}{\textbf{PeMS08$\to$PeMS04}} & \multicolumn{3}{c|}{\textbf{Chengdu$\to$Shenzhen}} & \multicolumn{3}{c}{\textbf{Shenzhen$\to$Chengdu}} \\
\cline{2-13}
& MAE & RMSE & MAPE & MAE & RMSE & MAPE & MAE & RMSE & MAPE & MAE & RMSE & MAPE \\
\hline\hline
w/o TU & \underline{18.955} & 30.344 & \underline{12.707} & 25.460 & 38.971 & 16.356 & 2.545 & \underline{3.467} & 10.787 & 3.464 & 5.184 & 14.817 \\
w/o SU & 19.163 & \underline{30.287} & 12.951 & \underline{24.870} & \underline{38.160} & \underline{15.602} & \textbf{2.433} & 3.521 & 10.662 & \underline{2.881} & \underline{4.387} & 12.440 \\
w/o STU & 21.808 & 34.594 & 19.320 & 30.439 & 48.587 & 25.562 & 2.591 & 3.508 & 11.260 & 3.470 & 5.277 & 15.316 \\
w/o Semi-implicit(Explicit Euler) & 22.363 & 34.750 & 20.845 & 31.660 & 51.306 & 28.361 & 2.608 & 3.572 & 11.949 & 3.530 & 5.388 & 15.379 \\
w/o Semi-implicit(RK4) & 22.350 & 34.745 & 20.786 & 31.726 & 51.534 & 28.575 & 2.620 & 3.585 & 11.960 & 3.538 & 5.399 & 15.469 \\
w/o MTS & 20.890 & 34.267 & 16.450 & 32.761 & 55.445 & 33.295 & 2.529 & 3.519 & \underline{10.630} & 2.949 & 4.435 & \underline{12.208} \\
MA-GLTC & \textbf{18.927} & \textbf{29.478} & \textbf{12.445} & \textbf{24.708} & \textbf{37.630} & \textbf{15.255} & \underline{2.435} & \textbf{3.462} & \textbf{10.620} & \textbf{2.845} & \textbf{4.074} & \textbf{12.047} \\
\hline\hline
\multirow{2}{*}{\textbf{Variant}} & \multicolumn{3}{c|}{\textbf{Chengdu$\to$PeMSBAY}} & \multicolumn{3}{c|}{\textbf{Shenzhen$\to$PeMSBAY}} & \multicolumn{3}{c|}{\textbf{PeMSBAY$\to$Chengdu}} & \multicolumn{3}{c}{\textbf{PeMSBAY$\to$Shenzhen}} \\
\cline{2-13}
& MAE & RMSE & MAPE & MAE & RMSE & MAPE & MAE & RMSE & MAPE & MAE & RMSE & MAPE \\
\hline\hline
w/o TU & 2.500 & 5.453 & 7.186 & 2.527 & 5.749 & 7.094 & 3.606 & 5.147 & 15.162 & 2.559 & 4.273 & 10.700 \\
w/o SU & 2.421 & \underline{5.450} & 7.187 & \underline{2.416} & \underline{5.455} & \underline{6.602} & \underline{3.014} & \underline{4.326} & 12.631 & 2.586 & 4.348 & 10.786 \\
w/o STU & 2.745 & 6.170 & 7.690 & 2.827 & 6.410 & 7.425 & 3.560 & 5.130 & 15.591 & \underline{2.466} & \underline{4.197} & \underline{10.659} \\
w/o Semi-implicit(Explicit Euler) & 3.216 & 6.242 & 7.850 & 2.810 & 6.463 & 7.477 & 3.625 & 5.169 & 15.602 & 2.508 & 4.257 & 10.760 \\
w/o Semi-implicit(RK4) & 3.233 & 6.272 & 7.969 & 2.816 & 6.499 & 7.486 & 3.689 & 5.177 & 15.648 & 2.520 & 4.261 & 10.780 \\
w/o MTS & \underline{2.386} & 5.603 & \underline{6.754} & 2.457 & 5.643 & 7.554 & 3.058 & 4.376 & \underline{12.580} & 2.561 & 4.285 & 10.667 \\
MA-GLTC & \textbf{2.365} & \textbf{5.416} & \textbf{6.680} & \textbf{2.404} & \textbf{5.395} & \textbf{6.524} & \textbf{2.942} & \textbf{4.320} & \textbf{12.524} & \textbf{2.459} & \textbf{3.754} & \textbf{9.931} \\
\hline\hline
\end{tabular}
\end{table*}

\begin{figure*}[htbp]
    \centering
    \begin{minipage}{0.48\textwidth}
        \centering
        \includegraphics[width=\textwidth]{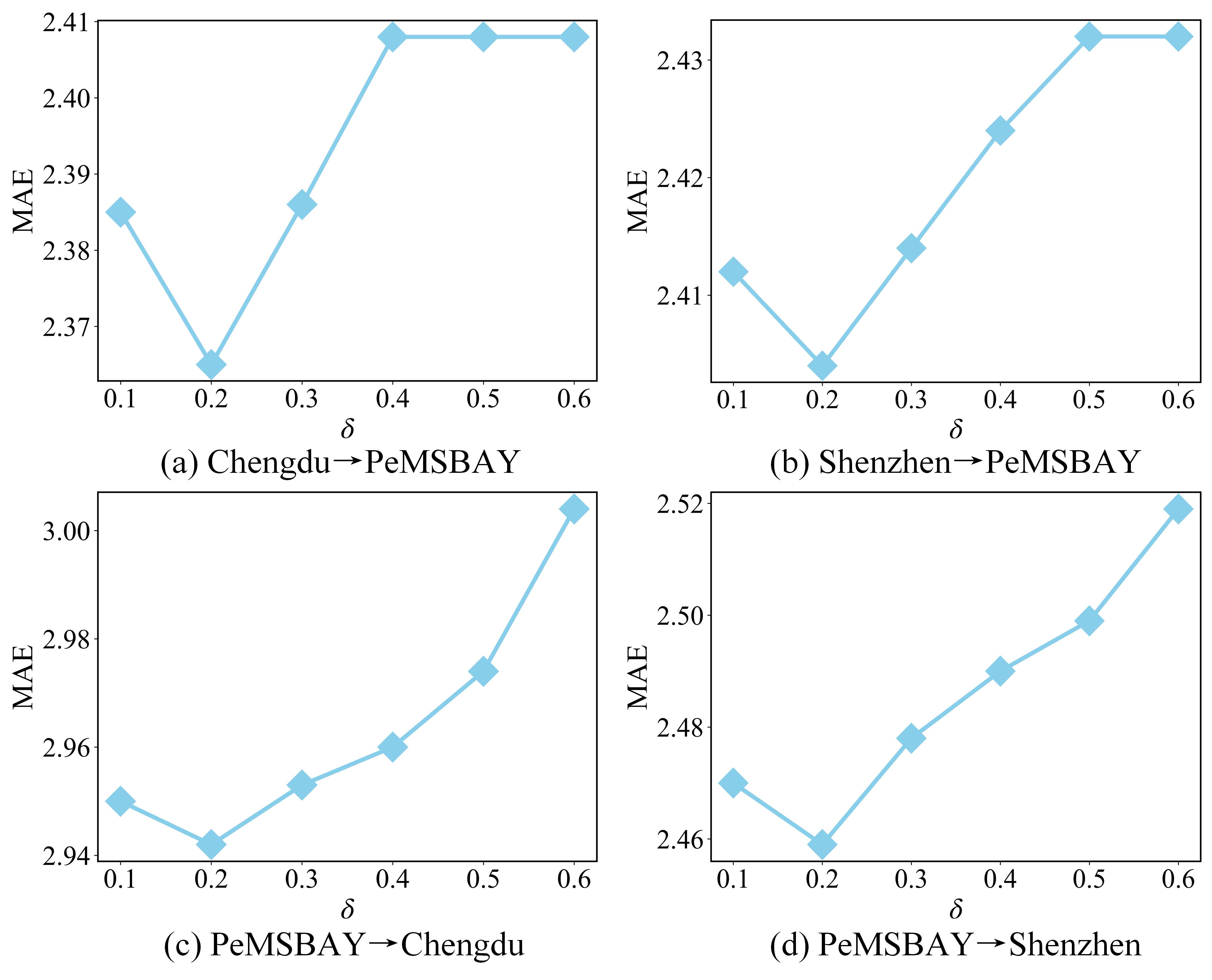}
        \caption{Prediction Error Analysis under Different Spatial-Temporal Unit Matching Thresholds $\delta$}
        \label{fig:zhexian}
    \end{minipage}
    \hfill
    \begin{minipage}{0.48\textwidth}
        \centering
        \includegraphics[width=\textwidth]{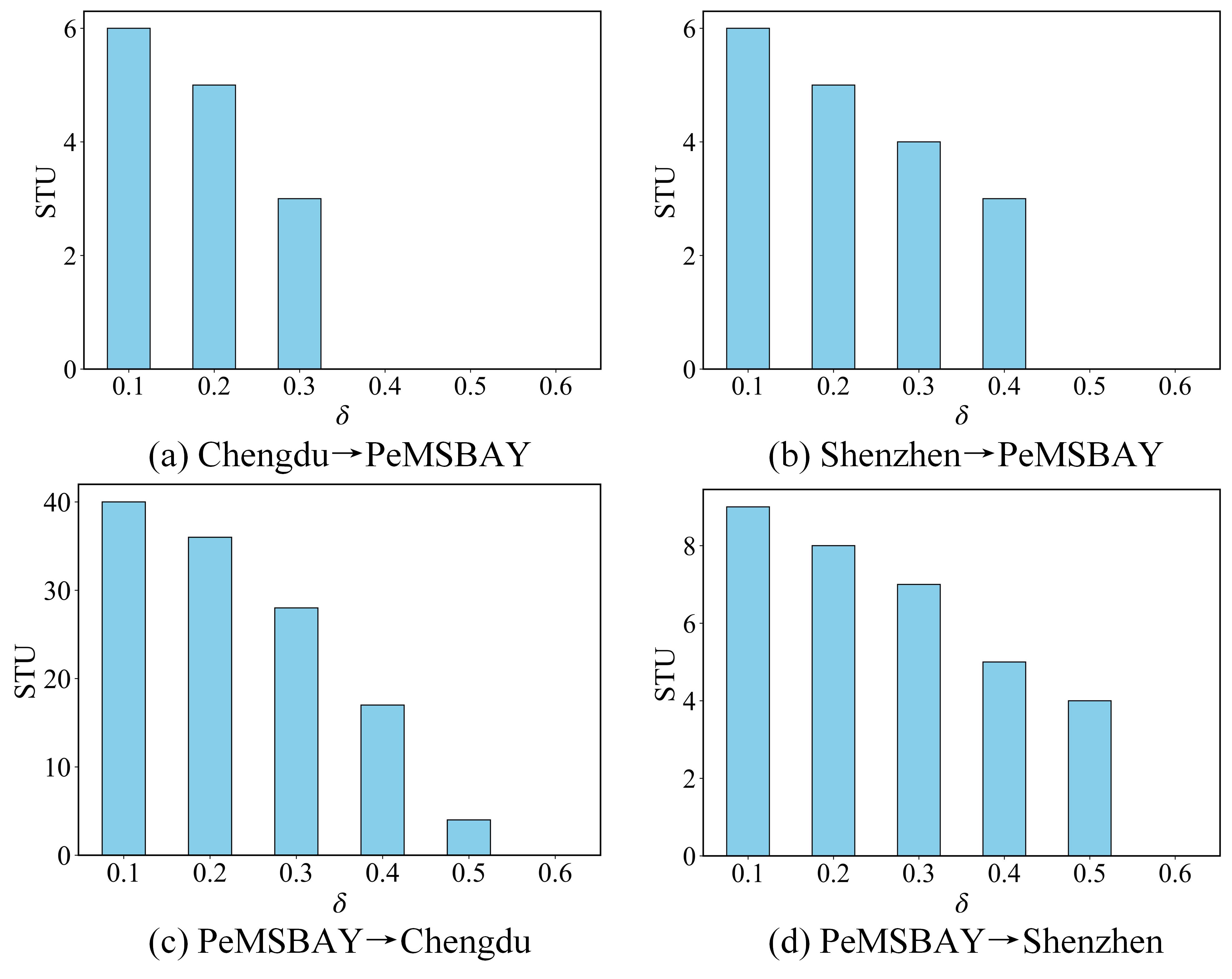}
        \caption{Variation of STU Count with Spatial-Temporal Unit Matching Thresholds}
        \label{fig:zhuzhuang}
    \end{minipage}
\end{figure*}

As shown in Table~\ref{tab:ablation}, removing any key component leads to performance degradation, confirming the effectiveness of the proposed design. Among these variants, w/o STU shows a clear drop, especially in short-term prediction, indicating that fine-grained spatio-temporal partitioning helps capture local traffic heterogeneity and improves unit-level transfer. Removing TU or SU also weakens performance, which further verifies that temporal and spatial partitions are both necessary for constructing effective transferable units. Replacing the semi-implicit solver degrades long-term prediction performance, suggesting that stable continuous-state updates are important for modeling nonlinear traffic evolution over extended horizons. In addition, w/o MTS performs worse under cross-domain settings with unseen target patterns, demonstrating that memory-based pattern storage and update are beneficial for preserving transferable source knowledge while adapting to target-domain distribution shifts. Overall, STU partitioning improves fine-grained transfer, GLTC provides stable continuous dynamics modeling, and MTS enhances adaptation to unseen traffic patterns. Their combination enables MA-GLTC to achieve robust cross-domain prediction under heterogeneous traffic conditions.

\subsubsection{Parameter Sensitivity Analysis of the STU Matching Threshold $\delta$}
We analyze the sensitivity of MA-GLTC to the CMD matching threshold $\delta$ on four transfer tasks with evident unseen patterns: Chengdu$\to$PeMSBAY, Shenzhen$\to$PeMSBAY, PeMSBAY$\to$Chengdu, and PeMSBAY$\to$Shenzhen. Figures~\ref{fig:zhexian} and~\ref{fig:zhuzhuang} show that $\delta$ affects both prediction accuracy and the number of identified unseen patterns. A small threshold, e.g., $\delta=0.1$, is strict and treats most target STUs as unseen, limiting the use of source-domain knowledge. In contrast, a large threshold may force dissimilar STUs to match existing patterns, introducing irrelevant transfer. The best overall performance is obtained at $\delta=0.2$, which balances source-pattern transfer and target-specific adaptation. Therefore, we set $\delta=0.2$ in all experiments. 

\begin{table*}[t]
\centering
\caption{Runtime and memory usage of MA-GLTC and learnable baselines on three networks.}
\label{tab:efficiency}
\scriptsize
\renewcommand{\arraystretch}{1}
\setlength{\tabcolsep}{9pt}
\begin{tabular}{l|c c c c c|c c c c c}
\hline\hline
\multirow{2}{*}{Method} & \multicolumn{5}{c|}{Time (s)} & \multicolumn{5}{c}{Memory (MB)} \\
\cline{2-11}
 & PeMS04 & PeMS08 & Chengdu & Shenzhen & PeMSBAY & PeMS04 & PeMS08 & Chengdu & Shenzhen & PeMSBAY \\
\hline
GRU & 7.436 & 4.235 & 12.889 & 8.383 & 21.035 & 9957.673 & 5583.915 & 8682.848 & 10340.360 & 5272.127 \\
TGCN & 8.387 & 4.762 & 8.657 & 10.642 & 28.370 & 643.336 & 344.256 & 539.737 & 659.619 & 659.670 \\
DCRNN & 8.892 & 5.173 & 9.866 & 12.012 & 30.393 & 162.994 & 96.098 & 212.290 & 253.931 & 394.796 \\
ASTTN & 16.209 & 10.360 & 30.216 & 38.420 & 52.630 & 15932.732 & 7346.562 & 16931.357 & 21927.370 & 17215.394 \\
IEEAFormer & 7.273 & 4.197 & 14.767 & 17.556 & 23.596 & 5988.529 & 2954.420 & 6282.643 & 8178.875 & 6460.624 \\
DASTNet & 4.674 & 4.736 & 7.688 & 5.705 & 4.121 & 1739.461 & 1738.367 & 2790.051 & 2807.750 & 3343.539 \\
STGFSL & 35.821 & 24.268 & 93.761 & 20.852 & 18.827 & 762.625 & 1375.915 & 2396.786 & 1992.700 & 2372.250 \\
STGP & 98.098 & 40.067 & 109.847 & 148.114 & 348.060 & 6629.462 & 3437.552 & 8643.737 & 13689.136 & 23154.249 \\
UniST & 163.880 & 139.990 & 365.450 & 670.190 & 924.710 & 90.669 & 79.558 & 86.561 & 78.450 & 79.431 \\
GDP & 1.920 & 1.825 & 1.956 & 1.883 & 4.910 & 550.621 & 552.410 & 547.047 & 549.902 & 551.531 \\
CGSTT & 6.576 & 3.665 & 11.915 & 12.220 & 16.711 & 3730.297 & 3158.097 & 4016.845 & 4264.474 & 4271.107 \\
MTPB & 8.396 & 4.273 & 15.760 & 13.222 & 17.712 & 590.635 & 812.266 & 1376.499 & 1202.324 & 1371.104 \\
MA-GLTC & 2.698 & 1.576 & 3.997 & 4.709 & 5.292 & 83.107 & 59.786 & 124.233 & 138.239 & 217.092 \\
\hline\hline
\end{tabular}
\end{table*}

\subsubsection{Computational efficiency and scalability analysis}
Table~\ref{tab:efficiency} shows that MA-GLTC achieves a strong efficiency–accuracy trade-off across five datasets. It trains much faster than attention-based or large-scale transfer models such as ASTTN, STGP, and UniST; for example, UniST takes 924.710 seconds per epoch on PeMSBAY, whereas MA-GLTC is substantially more efficient with better accuracy. MA-GLTC also has low memory cost, using only 59.786 MB on PeMS08 and 217.092 MB on PeMSBAY, while ASTTN exceeds 21 GB on Shenzhen. This efficiency comes from compact STU-level modeling and lightweight decoder adaptation. Although dense graph propagation remains the main scalability bottleneck, STU partitioning exploits traffic locality to reduce practical computation while preserving cross-domain prediction performance.

\section{CONCLUSION AND FUTURE WORK}
This paper proposes MA-GLTC, a cross-domain traffic prediction framework for data-scarce target regions. The framework constructs spatio-temporal units to support fine-grained cross-domain transfer, adopts GLTC to model graph-coupled traffic dynamics in continuous time, and introduces MTS to preserve source-domain knowledge while adapting to unseen target-domain patterns. Extensive experiments demonstrate that MA-GLTC achieves consistent improvements over representative inner-domain and cross-domain baselines, while ablation and sensitivity analyses further verify the effectiveness of STU partitioning, continuous-time dynamics modeling, and memory-based pattern adaptation. Despite these improvements, MA-GLTC still has several limitations. The semi-implicit solver may introduce errors around abrupt congestion transitions, and the MTS mechanism may bring additional memory overhead when the number of target-domain patterns increases. In future work, we will explore more adaptive continuous-time solvers and more efficient memory update strategies to further improve robustness and scalability on large-scale traffic networks.

\bibliographystyle{IEEEtran}
\bibliography{reference}

\end{document}